\documentclass[11pt]{article}
\usepackage[a4paper,margin=1in]{geometry}
\usepackage[T1]{fontenc}
\usepackage[utf8]{inputenc}
\usepackage{lmodern}
\usepackage{microtype}
\usepackage{amsmath,amssymb}
\usepackage{booktabs}
\usepackage{multirow}
\usepackage{xcolor}
\usepackage[numbers,sort&compress]{natbib}
\usepackage[colorlinks=true,linkcolor=blue!50!black,citecolor=blue!50!black,urlcolor=blue!50!black]{hyperref}
\usepackage{enumitem}
\usepackage{graphicx}
\usepackage{fancyvrb}
\usepackage{tikz}
\usepackage{orcidlink}
\usetikzlibrary{arrows.meta,positioning,shapes.geometric,fit,backgrounds,calc}
\definecolor{sxblue}{HTML}{2A78D6}
\definecolor{sxorange}{HTML}{EB6834}
\definecolor{sxink}{HTML}{52514E}
\graphicspath{{figures/}}
\hypersetup{
  pdftitle={System X: An LLM Decision Model That Replies from Its Own KV Cache},
  pdfauthor={Syed Usama Khalid Bukhari; Syed Safwan Khalid, Muhammad Aamir},
  pdfsubject={Decision models with joint decision and reply training},
  pdfkeywords={decision models, large language models, KV cache, calibration, customer support},
  bookmarksnumbered=true
}
\setlist{nosep,leftmargin=*}
\newcommand{\reply}{\texttt{REPLY\_WITH\_TEXT}}
\newcommand{\sysplus}{System~X+}
\newcommand{\sysfour}{System~X~v4}
\newcommand{\sysfoursmall}{System~X~v4-0.6B}

\title{System X: An LLM Decision Model That Replies from Its Own KV Cache}
\author{Syed Usama Khalid Bukhari\orcidlink{0000-0003-1581-8609}\\
\small Turing Logic LTD
\and
Syed Safwan Khalid\orcidlink{0000-0001-6248-0300}\\
\small 
\and
Muhammad Aamir\orcidlink{0000-0002-4999-7740}\\
\small University of Oxford}

\date{October 2026\\[0.5em]
\small Code: \url{https://github.com/Turinglogicuk/system-x}\\
Model: \url{https://huggingface.co/TuringLogic/system-x}}

\begin{document}
\maketitle

\begin{abstract}
Decision models return probabilities over supplied options, but a deployed agent may also need to reply in words. The present study introduces System~X, a large language model (LLM) decision model that adds a text-reply option to the LLM-as-Jev readout. When this option wins, or confidence is low, the same model continues from the decision's key--value (KV) cache, avoiding a repeated prompt prefill. Qwen3-0.6B has been fine-tuned with LoRA on Reason-9k and Bitext reply rows. JevBench accuracy shows no statistically significant change (55.4\% to 54.5\%, $p=0.87$), while expected calibration error falls from 0.292 to 0.169. Held-out Bitext routing accuracy reaches 98.3\%. Cache reuse saves 15--37\% of latency on an Apple M4 Pro, with token-identical replies in the FP32 demonstration cases. Scoring options on copies of the prompt cache matches full-sequence scoring to within $1.4\times10^{-6}$ in probability. Adding dialogue, industry and complaint data (System~X+) improves support-task results but lowers IFEval accuracy from 55.5\% to 45.0\%. System~X v4 uses Qwen3-1.7B, rebalanced data and teacher-rewritten reply targets. Against its base LLM, JevBench accuracy rises from 55.0\% to 61.0\% ($p=0.04$), with no statistically significant IFEval change ($p=0.71$). A matched latency test on unadapted Qwen3-1.7B saves 17--39\%. One LLM judge prefers v4 replies to an unadapted 1.7B second call (31 to 11 pairs, $p=0.003$); the other's preference is not significant. These reply comparisons also differ in prompts and generation limits. State conflicts and unsupported reply details remain failure modes. Further studies with several seeds, human raters and real support data are recommended.
\end{abstract}

\section{Introduction}

Many decisions inside software are choices among a known, short list: route a ticket, refund or escalate, approve or block. Decision models \citep{almeida2026jev} treat these as what they are. The caller provides a State and a list of options, and the model returns a probability for each option without generating free text. LLM-as-Jev \citep{li2026llmasjev} showed that an ordinary instruction-tuned LLM already does this well: pre-fill the assistant turn with \texttt{Best answer: [}, score each option's bracketed identifier \texttt{k]}, and normalise. That study reports targeted accuracy and calibration gains for small LLMs, with KL anchors reducing deterioration in its generation evaluations.

A decision model is only half of a customer-facing agent. When the right action is not offered, when the request is only a question, or when the model is unsure, the agent must reply in words. One design is a cascade: a decision model first, then an LLM call that reads the prompt again to write the reply. Without cache reuse, this costs a second prefill of the same context. Work on rubric judging reports overlapping errors between decision models and LLM judges \citep{rao2026rubric}; whether that finding transfers to customer support remains untested. System~X uses generation to communicate with the user, rather than to re-evaluate the decision.

The present study explores a simple question: \emph{can the decision model itself be the one that talks?} System~X combines one extra option with cache reuse:

\begin{enumerate}
\item \textbf{A reply option inside the decision readout.} \reply{} is listed with the other options and is scored using the same prompt prefill and batched suffix scoring. Its probability is learned jointly with the action probabilities, so ``I should say something'' is a decision like any other.
\item \textbf{Replies decoded from the decision's own KV cache.} If \reply{} wins, or the top probability is below a threshold $\tau$, the model continues from the prompt cache with \texttt{r] Reply:}, where $r$ is the reply option index, and writes one or two sentences. No token of the prompt is processed twice. In exact arithmetic this equals a second call on the same prompt, and in FP32 we observe token-identical replies.
\item \textbf{A training recipe and an honest evaluation.} The LLM-as-Jev objective has been combined with a reply language-model loss, a KL term on reply tokens and a Brier term, and trained on the paper's public data plus Bitext reply rows. Decisions, calibration, replies and latency have been measured, including where the model fails.
\end{enumerate}

Our contribution combines an explicit reply option, continuation from the decision prompt's cache, and joint decision and reply training in System~X (\S\ref{sec:related}). The method is architecture-preserving: it changes how an unmodified LLM is prompted, read out, decoded and trained, not the network itself.

The main aim of the present study is to evaluate whether a single small LLM can make calibrated customer-support decisions and, when needed, write the reply from the same key--value cache, as a fast adjunct tool for customer-support teams.

\section{Related work}\label{sec:related}

\paragraph{Decision models based on LLMs.}
Jev \citep{almeida2026jev} introduced ``system one'' decision models that return a categorical distribution over options; its weights and full training procedure are not publicly released. JevBench \citep{jevbench2026} evaluates such systems on accuracy, calibration and latency, with 231 public items. LLM-as-Jev \citep{li2026llmasjev} gives a training-free readout over bracketed identifiers and a fine-tuning objective with KL anchors; we reuse its decision-only prompt and readout and reproduce its Reason-9k data recipe. Our training adds reply and calibration losses, omits off-path KL anchors, and uses LoRA. Its published Qwen3-0.6B results therefore provide context, rather than a controlled ablation of our additions. AnyJev \citep{zhang2026anyjev} and Contrastive Language Models \citep{kwok2026clm} also produce decisions without generation; SemIf \citep{lee2026semif} reads option-letter logits. System~X explicitly represents a customer-facing text reply as an option in the decision list.

\paragraph{Falling back to generation.}
The \texttt{adaptive=True} mode of Tacit in the AnyJev code release \citep{anyjevrepo} routes uncertain decisions to additional reasoning. When the log-probability gap between the top two options is small, the same model is re-prompted with a fresh prefill, a different system prompt and thinking enabled, and returns a reasoning trace followed by a re-read label distribution. It differs from our design in three ways: it starts a new prefill rather than continuing the decision cache, its output is a revised decision distribution rather than a message to the user, and it has no reply option in the option list.

\paragraph{Cascades and confidence routing.}
JEV-as-a-Judge \citep{li2026jevjudge} accepts confident decision-model verdicts and escalates the rest to a reasoning judge. \citet{rao2026rubric} find that decision models and flash-tier LLM judges err in the same places, so such cascades mainly save cost. Our fallback uses the same LLM and has not been evaluated as an independent correctness check. We therefore use the fallback for \emph{communication}, asking or explaining when unsure, and not as a second opinion on correctness.

\paragraph{Calibration.}
We report expected calibration error (ECE, 15 equal-width bins) \citep{naeini2015ece,guo2017calibration} and the Brier score \citep{brier1950}, as LLM-as-Jev does. For $N$ evaluation rows, let $P_{ik}$ be the inference probability of option $k$, $g_i$ its gold index, $\hat k_i=\arg\max_k P_{ik}$, $c_i=\max_k P_{ik}$, and $a_i=\mathbb{1}[\hat k_i=g_i]$. We use
\begin{align}
\operatorname{ECE}&=\sum_{b:|B_b|>0}\frac{|B_b|}{N}
\left|\frac{1}{|B_b|}\sum_{i\in B_b}(a_i-c_i)\right|,\\
\operatorname{Brier}&=\frac{1}{N}\sum_{i=1}^{N}\sum_{k=1}^{K_i}
\bigl(P_{ik}-\mathbb{1}[k=g_i]\bigr)^2.
\end{align}
Here $B_b=\{i:(b-1)/15<c_i\le b/15\}$ for $b=1,\ldots,15$, with zero confidence included in $B_1$. The Brier score is summed over classes without dividing by $K_i$. Both metrics assess the distribution before the fallback rule.

\section{Method}

System~X does not modify the model architecture. It uses the unchanged Qwen3 transformer and tokenizer, with a LoRA adapter that can be merged into the weights after training; no layers or output heads are added. The contribution lies in the prompt, the readout, the decoding procedure and the training objective, so the method could be adapted to other instruction-tuned causal LLMs. The present implementation uses Hugging Face Transformers with explicit access to token logits and the KV cache; compatibility with other backbones and serving systems has not been evaluated.

\subsection{Readout (unchanged from LLM-as-Jev)}
Given a State $s$ and options $o_1,\dots,o_K$, the user message lists the State, then \texttt{[k] $o_k$} for each option, then \emph{``Select one option. Answer only with its bracketed numeric identifier.''} The chat template is applied with thinking off and no system message, and the assistant turn is pre-filled with \texttt{Best answer: [}. With thinking off, the Qwen3 template places an empty thinking block before this pre-fill (Appendix~\ref{app:prompt}). Call the resulting token sequence $x$. Option $k$ corresponds to the token suffix $c_k$ obtained by tokenizing the prompt followed by its numeric identifier and closing bracket, \texttt{k]}. We check that the prompt tokens remain an unchanged prefix. The closing bracket distinguishes identifiers such as \texttt{1]} and \texttt{12]}, avoiding a fixed single-token label limit; option count remains bounded by context length and memory. The option score and probability are
\begin{equation}
s_k=\sum_{t}\log p_\theta\!\left(c_{k,t}\mid x,\,c_{k,<t}\right),\qquad
P(k\mid x)=\frac{\exp s_k}{\sum_{j=1}^{K}\exp s_j}.\label{eq:readout}
\end{equation}
We compute all $s_k$ with one prefill of $x$, followed by teacher-forced scoring of the remaining suffix tokens on copies of the prompt cache. The first token of every suffix is scored from the prefill logits. The implementation processes at most 32 suffixes per batch, so larger option sets require multiple scoring batches. The original cache is left untouched. The repository includes a brute-force scoring check in \texttt{bench.py} to compare the cached and full-sequence paths. Rerunning it with the untrained Qwen3-0.6B in FP32 on an NVIDIA RTX~PRO~4000, the two paths agree to within $3\times10^{-8}$ in probability on the first eight JevBench items, and to within $1.4\times10^{-6}$ on a 40-option prompt whose two- and three-token suffixes are scored in two batches.

\subsection{The reply option}\label{sec:replyopt}
When \reply{} is among the options, it is listed like any other option, with the description \emph{``Write a 1-2 sentence reply (answer a question, or explain what the customer can do)''}. One sentence is appended to the instruction: \emph{``If you select \reply{}, write a 1-2 sentence reply to the customer after the identifier.''} Without the reply option, the prompt is exactly the LLM-as-Jev prompt. All training rows use the description above. The engine's own option catalogue, which supplies the options of the hand-written set, the demonstrations, the latency test and the web console, describes \reply{} differently: \emph{``Write a 1-2 sentence reply (answer a question, or ask for missing details)''}.

\subsection{Deciding and replying from one cache}
\begin{figure}[t]
\centering
\begin{tikzpicture}[
  font=\footnotesize, >={Stealth[length=2.2mm]},
  sxbox/.style={draw=sxink!70, rounded corners=2pt, fill=white, align=center, inner sep=4pt, minimum height=11mm},
  cache/.style={draw=sxblue, fill=sxblue!10, cylinder, shape border rotate=90, aspect=0.22, minimum width=16mm, minimum height=12mm, align=center, inner sep=2pt},
  outbox/.style={sxbox, draw=sxblue, fill=sxblue!8},
  flow/.style={->, draw=sxink, line width=0.7pt},
  lbl/.style={font=\scriptsize, text=sxink}]
\node[sxbox, text width=30mm] (x) at (1.6,0) {\textbf{Prompt $x$}\\[-1pt]{\scriptsize State $+$ options\\[-2pt]$o_1,\dots,o_K$ $+$ \texttt{Best answer: [}}};
\node[sxbox, text width=18mm] (pre) at (4.3,0) {\textbf{1 prefill}\\[-1pt]{\scriptsize reads $x$ once}};
\node[cache] (kv) at (6.5,0) {\scriptsize KV cache\\[-2pt]\scriptsize of $x$};
\node[sxbox, text width=40mm] (score) at (10.0,0) {\textbf{Batched suffix scoring}\\[-1pt]{\scriptsize \texttt{1]}, \texttt{2]}, \dots, \texttt{K]} on cache copies\\[-2pt]$\Rightarrow P(1\,|\,x),\dots,P(K\,|\,x)$}};
\node[diamond, draw=sxink!70, aspect=2.4, inner sep=1.5pt, align=center, font=\scriptsize] (gate) at (10.2,-2.9) {$\hat k=r$ or\\[-2pt]$P(\hat k\,|\,x)<\tau$?};
\node[outbox, text width=24mm] (act) at (14.2,-2.9) {\textbf{Return action $\hat k$}\\[-1pt]{\scriptsize no text generated}};
\node[outbox, text width=40mm] (reply) at (5.8,-2.9) {\textbf{Reply from the same cache}\\[-1pt]{\scriptsize append \texttt{r] Reply:}, decode greedily\\[-2pt]$\Rightarrow$ 1--2 sentence reply}};
\node[sxbox, draw=sxorange, fill=sxorange!8, text width=30mm] (two) at (1.6,-2.9) {\textbf{Two-call baseline}\\[-1pt]{\scriptsize re-reads $x$ in a second\\[-2pt]prefill before replying}};
\draw[flow] (x) -- (pre);
\draw[flow] (pre) -- (kv);
\draw[flow] (kv) -- (score);
\draw[flow] (score) -- (gate);
\draw[flow] (gate) -- node[lbl, above] {no} (act);
\draw[flow] (gate) -- node[lbl, above] {yes} (reply);
\draw[->, draw=sxblue, line width=1.1pt] (kv.south) -- node[lbl, left, text=sxblue] {cache reused} (kv.south |- reply.north);
\draw[->, draw=sxorange, line width=0.8pt, dashed] (x.south) -- node[lbl, right, text=sxorange!80!black] {2nd prefill} (two.north);
\end{tikzpicture}
\caption{System~X decides and replies from one prefill. The prompt is read once; every option's suffix is scored on copies of its KV cache, which stays untouched. A confident action returns without text. If \reply{} (index $r$) wins or the top probability falls below $\tau$, decoding continues from the same cache, so only \texttt{r] Reply:} and the reply tokens are processed. A two-call pipeline instead re-reads $x$ before replying (orange, dashed); that second prefill is the latency the method saves.}
\label{fig:pipeline}
\end{figure}
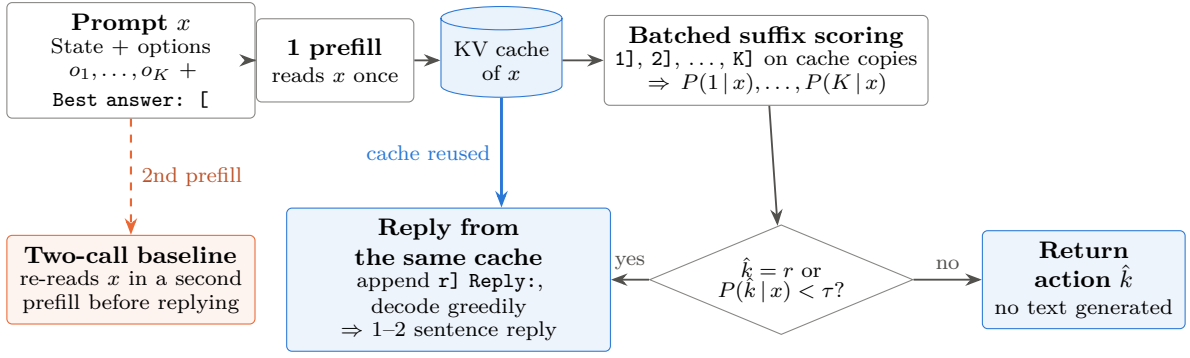

When \reply{} is present, let $r$ be its index and $\hat k=\arg\max_k P(k\mid x)$. The engine returns the decision $\hat k$ with no text if $\hat k\neq r$ and $P(\hat k\mid x)\ge\tau$. Otherwise, let $d_r$ be the tokenized continuation containing the numeric reply-option index followed by \texttt{] Reply:}. The reply is decoded one token at a time:
\begin{equation}
y_t=\arg\max_{v\in\mathcal V}p_\theta\!\left(v\mid x,d_r,y_{<t}\right),\qquad t=1,2,\ldots,
\end{equation}
where $\mathcal V$ is the LLM vocabulary and ties are resolved consistently. Generation stops at an end token or the configured generation limit (48 tokens in the reply evaluation). Without \reply{} in the option list, System~X returns the argmax decision without generating text. Otherwise, two triggers lead to a reply: \reply{} wins, or no option is confident enough (a \emph{low-confidence fallback}). Figure~\ref{fig:pipeline} summarises the procedure.

For fixed model weights and inference settings, the prompt KV cache is determined by $x$. Scoring the suffixes uses copies, so the cache still holds exactly $x$, and decoding $y$ only has to process the few tokens of \texttt{r] Reply:} and then the reply itself. A two-call pipeline would prefill $x$ a second time. The saving is therefore one prefill of the prompt, and it grows with prompt length. With identical weights, tokenization, positions, attention masks and greedy decoding, the two paths are equivalent in exact arithmetic. In FP32 we observe identical tokens on the demonstration cases; in bf16, rounding differences can occasionally flip a greedy choice.

\subsection{Training objective}
For a row with gold option $g$, let $L=|c_g|$ and let $n_t$ denote the prompt $x$ together with the gold suffix prefix $c_{g,<t}$ at its corresponding trie node, $\mathcal{A}(n)$ the legal next tokens at node $n$, and $m(n)=\sum_{a\in\mathcal{A}(n)}p_\theta(a\mid n)$ the legal mass. Following LLM-as-Jev, the decision loss is the tree-factorized listwise loss
\begin{equation}
\mathcal{L}_{\text{tree}}=-\sum_{t=1}^{L}\log\frac{p_\theta(c_{g,t}\mid n_t)}{m(n_t)}.\label{eq:tree}
\end{equation}
This loss is the negative log-probability of the gold leaf under local normalization at each trie node. It is generally different from $-\log P(g\mid x)$ in Eq.~\eqref{eq:readout}, which normalizes complete suffix likelihoods. At a node with only one legal next token, the contribution to Eq.~\eqref{eq:tree} is zero. We add KL anchors to the frozen base model $p_0$, using reference statistics cached once (top-64 tokens plus a remainder bucket): $\mathcal{L}_{\text{mass}}$, a binary KL on the legal mass at gold-path nodes; $\mathcal{L}_{\text{out}}$, a KL on the renormalised illegal-token distribution at those nodes; and $\mathcal{L}_{\text{pos}}$, a KL over the top-64 tokens and remainder bucket at the start of the assistant turn and, on decision rows, right after the closing bracket. The mass and illegal-token terms each use weight $0.5/L$ per gold-path node; the positional terms each use weight $0.5$ per site. All KL terms use the direction $D_{\mathrm{KL}}(p_0\Vert p_\theta)$. We add three terms:
\begin{itemize}
\item a Brier term $\mathcal{L}_{\text{Brier}}=\sum_k\left(q_k-\mathbb{1}[k=g]\right)^2$ on the first-token distribution $q_k=p_\theta(c_{k,1}\mid x)/\sum_j p_\theta(c_{j,1}\mid x)$, computed for rows with $K\le9$, where each option's first token identifies it uniquely. This is a calibration proxy: unlike the inference distribution $P(k\mid x)$, it omits the probability of the remaining suffix tokens;
\item for reply rows, a mean language-model loss $\mathcal{L}_{\text{LM}}$ on the reply text $y$ and the end-of-turn token, conditioned on the fixed prefix \texttt{r] Reply:};
\item a mean KL term $\mathcal{L}_{\text{rKL}}$ to $p_0$ on the same reply-token positions, which regularises the reply distribution toward the base LLM.
\end{itemize}
The full loss is
\begin{equation}
\mathcal{L}=\mathcal{L}_{\text{tree}}+\lambda\left(\mathcal{L}_{\text{mass}}+\mathcal{L}_{\text{out}}+\mathcal{L}_{\text{pos}}\right)+\gamma\,\mathcal{L}_{\text{Brier}}+\beta\,\mathcal{L}_{\text{LM}}+\lambda_r\,\mathcal{L}_{\text{rKL}},
\end{equation}
with $\lambda=1$, $\gamma=1$, $\beta=1$ and $\lambda_r=0.5$. The Brier term is zero when $K>9$; the LM and reply-KL terms are zero for rows without a reply target. Reference probabilities are stored in FP16, and the implementation clamps probabilities to avoid logarithms of zero, so the KL computations are numerical approximations to the expressions above. Unlike LLM-as-Jev, we do not anchor off-path trie nodes.

\subsection{Data}
\paragraph{Decision rows (9{,}000).} We use the dataset composition and sample counts of the LLM-as-Jev Reason-9k mixture: 1{,}000 rows each from CLINC150 \citep{larson2019clinc}, MASSIVE en-US \citep{fitzgerald2023massive}, Bitext customer support \citep{bitext2024}, ReClor \citep{yu2020reclor}, LogiQA~2.0 \citep{liu2023logiqa2}, CosmosQA \citep{huang2019cosmosqa}, PIQA \citep{bisk2020piqa}, $\alpha$NLI \citep{bhagavatula2020anli} and ProofWriter \citep{tafjord2021proofwriter}. As in the paper, CosmosQA's ``none of the above'' answers are upsampled to 16\%, ProofWriter is balanced, and every option list is shuffled. These rows have no reply option.

\paragraph{Reply-task rows (3{,}000, ours).} We build these from Bitext's hybrid synthetic customer-support dataset \citep{bitext2024}, whose messages are generated from natural-text seeds with linguistic curation. Seventeen Bitext intents are mapped to 15 actions (for example, the intent \texttt{cancel\_order} becomes the action \texttt{CANCEL\_ORDER}, and the intent \texttt{contact\_human\_agent} becomes \texttt{ESCALATE\_HUMAN}). Each row offers 2--5 distractor actions plus \reply{}, in shuffled order. The rows come in three kinds:
\begin{itemize}
\item \textbf{action (50\%):} the message's action is offered, and it is the label;
\item \textbf{info (35\%):} an information intent, such as a refund-policy question; the label is \reply{};
\item \textbf{not offered (15\%):} an action intent whose action is missing from the list; the label is \reply{}.
\end{itemize}
Reply targets are one or two informative sentences extracted from Bitext's response: we drop openers, empathy filler and list fragments, then keep up to two sentences that pass a content-cue filter in their original order. A second sentence is kept only if the pair stays within 45 words, and targets outside 8--55 words are discarded. The State contains only the customer message. Distinct messages are split 90/10 by normalised text, before sampling either Bitext training component. We then sample 3{,}000 reply-task training rows and a separate 300-row held-out set with the same 50/35/15 mix. No normalised held-out message occurs in either Bitext training component.

\section{Experimental setup}

\paragraph{Model and training.} System~X uses the Qwen3-0.6B LLM \citep{yang2025qwen3} with LoRA \citep{hu2022lora}: $r=16$, $\alpha=32$, on all attention and MLP projections (10.1M trainable parameters). We train one epoch (375 steps of 32 rows) with AdamW, learning rate $10^{-4}$, 10\% linear warm-up and linear decay to 10\% of the peak, gradient clipping at 1.0, in FP32 with gradient checkpointing. On an Apple M4 Pro (24\,GB, \texttt{mps}) training takes 90 minutes, after an 18-minute pass to cache the base model's reference statistics. We use one training run (PyTorch seed 20260922 and example-order seed 0). AdamW uses weight decay 0.01, and LoRA dropout is zero. Each 32-row step is split into micro-batches whose padded length times row count stays within a token budget (4{,}000 tokens here, 16{,}000 for the later runs). Every row's loss is divided by the step's row count, so the budget changes memory use but not the objective. The frozen model's reference statistics are computed in FP32. The dataset builders use random seed 20261015 for System~X and 20261016 for \sysplus{} and \sysfour{}.

\paragraph{Evaluation sets.}
\begin{itemize}
\item \textbf{JevBench public} \citep{jevbench2026}: 231 items, run in FP32 as in LLM-as-Jev. These are the same 231 items the paper reports on.
\item \textbf{Banking77} \citep{casanueva2020banking77}: the test set, 3{,}080 rows with 77 options each, run in bf16.
\item \textbf{Held-out Bitext reply task}: the 300 rows described above.
\item \textbf{Hand-written}: 97 customer-support cases we wrote ourselves, with different wording, option descriptions and domain details from any training data. Three of its seven action names (\texttt{ESCALATE\_HUMAN}, \texttt{TRACK\_ORDER}, \texttt{UPDATE\_ADDRESS}) also occur among the System~X training actions, with different descriptions. They include 56 action cases (fraud lock, refund, escalation, \dots), 8 state conflicts (e.g.\ an address change on an already-shipped order, where the right answer is to reply), 14 information questions, 10 vague messages and 9 cases where the right action is not offered. Its \reply{} description is the engine catalogue's, not the training one (\S\ref{sec:replyopt}).
\end{itemize}
Unless stated otherwise, accuracy and ECE describe the argmax decision before fallback. We also report accuracy after applying the fallback at $\tau=0.5$; the engine's default, used in the demonstrations of Appendix~\ref{app:examples}, is $\tau=0.65$. Both reply evaluation sets use bf16. Reported paired $p$-values use exact two-sided tests on discordant outcomes. These comparisons are exploratory and are not adjusted for multiple testing; a nonsignificant result does not establish equivalence.

\paragraph{Latency protocol.} We compare System~X with cache reuse against a two-call pipeline that runs the same decision and then re-reads the whole prompt to write the reply. Both use the same \texttt{generate()} loop and write exactly 32 reply tokens (median of 5 runs, bf16, \texttt{mps}). Longer prompts are built by appending a synthetic account history to the State until its JSON reaches about 200, 500, 1{,}000, 2{,}000 and 4{,}000 tokens; the full prompts are 367 to 4{,}161 tokens long. Each prompt offers four catalogue options including \reply{}, and a threshold above 1 forces the reply path. The two-call time covers a decision call that writes no reply plus a reply call that re-reads the prompt, timed after one warm-up call. An earlier comparison that used a hand-written decoding loop for one path and \texttt{generate()} for the other was misleading, because loop overhead dominated; we report only the matched comparison.

\section{Results}

\subsection{Decision accuracy and calibration after joint training}

\begin{table}[t]
\centering
\caption{Decisions on public benchmarks with Qwen3-0.6B. ``Paper'' columns are from LLM-as-Jev \citep{li2026llmasjev}. Their Reason-9k row uses full fine-tuning. Their best JevBench row uses LoRA on Reason+Long-12k with $\lambda=0.01$. System~X uses LoRA on Reason-9k plus reply-task rows; these are different training conditions.}
\label{tab:bench}
\small
\begin{tabular}{lccccc}
\toprule
& \multicolumn{2}{c}{Ours} & \multicolumn{3}{c}{Paper} \\
\cmidrule(lr){2-3}\cmidrule(lr){4-6}
& Base LLM & System~X & Base LLM & Reason-9k & Best JevBench run \\
\midrule
JevBench accuracy (\%) & 55.4 & 54.5 & 56.7 & 55.0 & 61.0 \\
JevBench ECE & 0.292 & \textbf{0.169} & 0.278 & 0.118 & 0.121 \\
JevBench Brier & 0.674 & \textbf{0.575} & -- & -- & -- \\
Banking77 accuracy (\%) & 18.5 & \textbf{53.5} & 22.2 & 60.4 & 59.0 \\
Banking77 ECE & 0.460 & \textbf{0.079} & -- & -- & -- \\
\bottomrule
\end{tabular}
\end{table}

The results are shown in Table~\ref{tab:bench}. Our untrained model scores 55.4\% on JevBench against the paper's 56.7\%, a difference of three items that does not establish exact reproduction. After training with the reply task, the difference in JevBench accuracy is not statistically significant: 18 items are won and 20 lost, exact two-sided sign test $p=0.87$. This test does not establish equivalence or isolate the effect of adding the reply task; that requires a matched decision-only training run. With only 231 items, this evaluation has limited power to detect small effects. The analysis of calibration revealed a marked improvement, with ECE falling from 0.292 to 0.169 and Brier from 0.674 to 0.575, though it remains above the paper's 0.118. Banking77 accuracy rises by 35 percentage points to 53.5\%, compared with the paper's 60.4\%; our ECE is 0.079. Our untrained Banking77 score (18.5\%) is also below the paper's 22.2\%. We ran Banking77 in bf16 and did not investigate this gap. By tier, the easy JevBench items go from 45/48 to 48/48 and the hard items from 36/111 to 38/111, while the ``original'' tier falls from 47/72 to 40/72.

\subsection{Reply routing improves on held-out Bitext messages}

\begin{table}[t]
\centering
\caption{The reply task. Held-out: 300 Bitext messages never seen in training. Hand-written: 97 out-of-distribution cases. Accuracy by kind is measured before fallback. ROUGE-L uses only gold-reply rows on which a reply was generated (150 base-LLM rows and 147 System~X rows), so the evaluated subsets differ.}
\label{tab:reply}
\small
\begin{tabular}{llcc}
\toprule
Set & Metric & Base LLM & System~X \\
\midrule
\multirow{6}{*}{Held-out Bitext}
& accuracy (\%) & 58.7 & \textbf{98.3} \\
& \quad action / info / not offered & 17.3 / 100 / 100 & 98.7 / 98.1 / 97.8 \\
& ECE & 0.389 & \textbf{0.020} \\
& Brier & 0.778 & \textbf{0.025} \\
& reply ROUGE-L F1 & 0.099 & \textbf{0.231} \\
& share answered with a reply (\%) & 91.0 & 49.3 \\
\midrule
\multirow{6}{*}{Hand-written}
& accuracy (\%) & 47.4 & \textbf{78.4} \\
& \quad actions (56) & 10.7 & 71.4 \\
& \quad info / vague / not offered (33) & 100 & 100 \\
& \quad state conflicts (8) & 87.5 & 37.5 \\
& accuracy after fallback (\%) & 47.4 & 76.3 \\
& ECE & 0.491 & \textbf{0.124} \\
\bottomrule
\end{tabular}
\end{table}

The results in Table~\ref{tab:reply} revealed that, untrained, the model almost always picks the reply option: 91\% of held-out rows and 93\% of hand-written rows. This makes it right on every question and wrong on most actions. After training, it separates the three held-out kinds at 97.8--98.7\% accuracy with an ECE of 0.020. Its 49.3\% reply share is close to the 50\% gold reply share. On the out-of-distribution hand-written set, accuracy rises from 47.4\% to 78.4\%, but this average hides the two failure modes in \S\ref{sec:errors}.

The low-confidence fallback reduces accuracy on this set. With $\tau=0.5$, hand-written accuracy including the fallback is 76.3\%, below the argmax accuracy. Action accuracy falls from 71.4\% (40/56) before fallback to 67.9\% (38/56) after it. Only one of the 21 errors has a top probability under 0.5, and 13 have 0.80 or more. The fallback does not catch confident mistakes.

\subsection{One prefill saved}

\begin{table}[t]
\centering
\caption{End-to-end latency of decision plus a 32-token reply, Apple M4 Pro, bf16, median of 5 runs. Both paths use the same \texttt{generate()} loop; ``two calls'' re-reads the prompt for the reply.}
\label{tab:latency}
\small
\begin{tabular}{rrrrr}
\toprule
Prompt tokens & Decision only & System~X & Two calls & Saved \\
\midrule
367   & 134 ms   & 414 ms   & 488 ms   & 74 ms (15\%) \\
679   & 215 ms   & 508 ms   & 639 ms   & 131 ms (21\%) \\
1{,}159 & 350 ms & 671 ms   & 901 ms   & 230 ms (26\%) \\
2{,}159 & 654 ms & 1{,}019 ms & 1{,}494 ms & 476 ms (32\%) \\
4{,}161 & 1{,}386 ms & 1{,}824 ms & 2{,}891 ms & 1{,}067 ms (37\%) \\
\bottomrule
\end{tabular}
\end{table}

\begin{figure}[t]
\centering
\includegraphics[width=\linewidth]{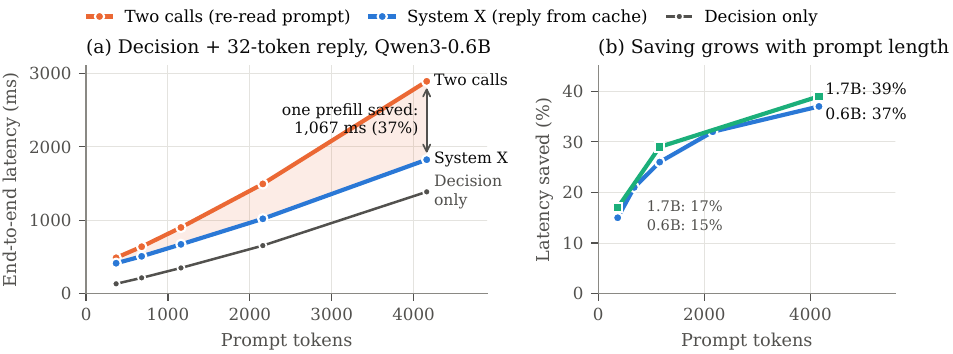}
\caption{Latency of cache reuse (Apple M4 Pro, bf16, median of 5 runs). (a) End-to-end time for a decision plus a 32-token reply with Qwen3-0.6B (Table~\ref{tab:latency}); the shaded gap is the second prefill that System~X avoids. (b) Relative saving for Qwen3-0.6B and for the untrained Qwen3-1.7B backbone (\S\ref{sec:v4}), at the prompt lengths measured for each.}
\label{fig:latency}
\end{figure}

Table~\ref{tab:latency} revealed a latency reduction from avoiding a repeated prefill, which grows with prompt length: from 15\% at 367 tokens to 37\% at 4{,}161. Figure~\ref{fig:latency} plots these times together with the 1.7B measurements of \S\ref{sec:v4}.

 On the held-out Bitext messages, whose prompts are short and whose replies have their natural length, the medians over the 29 reply-generating cases among the first 60 rows are 292\,ms for the one-model path and 325\,ms for the two-call path. A decision alone takes a median of 80\,ms at these lengths. This comparison uses the same LLM for both paths and does not measure memory use or the quality of a separate reply model. In FP32, the cached reply is token-for-token identical to the re-read reply on all our demonstration cases.

\subsection{Error analysis}\label{sec:errors}

\paragraph{The reply option attracts unfamiliar actions.}
Of the 21 hand-written errors, 14 send an action case to \reply{}:
\begin{itemize}
\item six safety, legal or complaint escalations (``The e-bike battery swelled up and started hissing in my hallway'', ``Expect to hear from my attorney if this isn't resolved by Friday'');
\item four account-takeover cases that should be \texttt{BLOCK\_FRAUD};
\item three refunds;
\item one address change.
\end{itemize}
These go to \reply{} at 0.54--0.96 confidence, 8 of the 14 at 0.80 or more. Fraud is absent from the 15 actions in our reply data. Escalation and refunds are present (\texttt{ESCALATE\_HUMAN}, \texttt{START\_REFUND}), but differently: training escalations occur only when the customer asks for a human, never for safety or legal threats, and the hand-written set describes its options differently and names most of them differently: \texttt{ESCALATE\_HUMAN} keeps its training name, while \texttt{AUTO\_REFUND} replaces \texttt{START\_REFUND}. These errors suggest that unfamiliar action descriptions can attract the reply option with high confidence. This mirrors the ``none of the above'' over-selection that LLM-as-Jev reports for single-domain data. Whether broader action data reduces this failure requires a controlled experiment.

\paragraph{Decisions can contradict the account state.}
The State in our reply data holds only the customer message, so this component of training provides no supervision for conditioning decisions on account facts. Five of the eight state conflicts are confident actions that the State rules out:
\begin{itemize}
\item \texttt{UPDATE\_ADDRESS} on an order that is \texttt{shipped} or \texttt{out\_for\_delivery}, at 0.86 and 0.99 confidence;
\item \texttt{CANCEL\_SUBSCRIPTION} when \texttt{subscription\_active} is false, three times, at 0.99--1.00.
\end{itemize}
The untrained model was right on seven of these eight, but only because it replies to almost everything.

\paragraph{Reply quality.}
On held-out rows, replies are short (median 18 words) and 96.6\% stop before the 48-token cap, but they inherit Bitext's style. 19 of 148 replies (13\%) contain the placeholder marker \texttt{\{\{} such as \texttt{\{\{Order Number\}\}}. The model also sometimes echoes the instruction (``Write a 1-2 sentence reply to the customer.''). We count an echo when the output contains the substring \texttt{1-2 sentence}. The untrained model does this in 66\% of its replies (179/273); after training the rate falls to 20\% on held-out rows (30/148), but it is 40\% on the hand-written set (21/53). A ROUGE-L F1 \citep{lin2004rouge} of 0.231 against Bitext's reply should be read with these issues in mind. Appendix~\ref{app:examples} shows examples.

\section{Broader support data: \sysplus}\label{sec:plus}

In the repository, System~X corresponds to v2 and \sysplus{} to v3; both use Qwen3-0.6B. The two failure modes above suggest limitations in the reply data: it covers 15 actions and no account state. We therefore trained a second model, \sysplus, with three further public sources added to the System~X mixture. The System~X rows are kept unchanged, while data, training duration and training hardware change together; this is not a controlled data-only ablation.

\subsection{Data and training}
\begin{itemize}
\item \textbf{ABCD} \citep{chen2021abcd} (5{,}000 rows): next-step decisions in crowdworker customer-support dialogues from the Action-Based Conversations Dataset, collected around prescribed scenarios for a fictitious company. The State holds the last ten turns and the actions taken so far. Account and order facts enter the State only after the lookup action that reveals them in ABCD (\texttt{pull-up-account}, \texttt{verify-identity} or \texttt{validate-purchase}). The label is the agent's next action, mapped to 13 renamed actions, or \reply{} with the agent's next message as the target. The mix is 45\% actions and 55\% replies; rows start after the customer's first message, and consecutive agent messages are merged. Merged replies of 4--50 words are kept, and each row offers 3--6 distractor actions plus \reply{}.
\item \textbf{Bitext verticals} \citep{bitextverticals2024} (5{,}000 rows): ten further Bitext industry datasets (retail banking, insurance, telco, retail e-commerce, travel, mortgages and loans, wealth management, hospitality, events and ticketing, media), with the same action / info / not-offered construction and 50/35/15 mix as the System~X reply rows. We add synthetic company and customer facts to the State (website, support phone, hours, order number, card) and fill Bitext's template slots in the reply targets from those facts, so targets that use those slots can be grounded in the supplied State. Rows whose targets contain a slot we cannot fill are dropped.
\item \textbf{CFPB complaints} \citep{cfpb2026} (1{,}496 rows): the first sentences (at most 70 words) of real consumer complaints to the US Consumer Financial Protection Bureau. Fraud, unauthorized-transaction and identity-theft complaints (40\%) are mapped to a fraud action; seven non-fraud issues (disputes, fees, card problems, account closure, payments, hardship, account access) serve as negatives. Because complaint categories are noisy, a row is kept only if its message mentions a cue for its class. Dates and redaction marks are normalised, and each row offers 2--4 distractor actions. \reply{} is offered as a distractor in half of the rows and is never the label.
\end{itemize}
All new option names and descriptions differ from the hand-written test set's; the build script checks that none of that set's option names or descriptions appear in the new rows. The full mixture has 23{,}496 rows; customer-support rows rise from 33\% of the System~X mixture to 66\%, and reasoning rows fall from 50\% to 26\%. We later found that 113 vertical reply targets (0.48\% of all 23{,}496 training rows; 4.52\% of the 2{,}500 vertical reply targets) contain a doubled word (``app app'') from slot filling.

Training uses the System~X settings: one epoch, now 735 optimizer steps with up to 32 rows per step (8 in the final step), the same seeds, FP32 weights and a token budget of 16{,}000 per micro-batch. It ran on one NVIDIA RTX~5090 in 32 minutes. Held-out sets: 300 rows from ABCD's official test split, 300 Bitext-vertical rows split by normalised message within each industry (150 action, 100 info and 50 not-offered rows after per-industry quota rounding), and 199 CFPB complaints split by complaint ID. All decision results below were measured on the Apple M4 Pro, as for System~X.

\subsection{Decision results}

\begin{table}[t]
\centering
\caption{Decisions of the base LLM, System~X and \sysplus{} (argmax, before fallback). JevBench in FP32; all other sets in bf16. ABCD, Bitext verticals and CFPB are held out from \sysplus{}'s training sources and absent from the other two fine-tuning mixtures; base-model pretraining overlap was not assessed.}
\label{tab:plus}
\small
\setlength{\tabcolsep}{4pt}
\begin{tabular}{lcccccc}
\toprule
& \multicolumn{3}{c}{Accuracy (\%)} & \multicolumn{3}{c}{ECE} \\
\cmidrule(lr){2-4}\cmidrule(lr){5-7}
Set & Base LLM & System~X & \sysplus{} & Base LLM & System~X & \sysplus{} \\
\midrule
JevBench (231) & 55.4 & 54.5 & 52.4 & 0.292 & \textbf{0.169} & 0.201 \\
Banking77 (3{,}080) & 18.5 & 53.5 & \textbf{57.2} & 0.460 & 0.079 & 0.081 \\
Held-out Bitext (300) & 58.7 & 98.3 & 99.0 & 0.389 & 0.020 & 0.019 \\
Hand-written (97) & 47.4 & \textbf{78.4} & 76.3 & 0.491 & \textbf{0.124} & 0.139 \\
ABCD next step (300) & 52.3 & 51.7 & \textbf{88.7} & 0.420 & 0.270 & \textbf{0.037} \\
\quad actions (135) & 7.4 & 36.3 & \textbf{83.0} & & & \\
Bitext verticals (300) & 58.3 & 86.3 & \textbf{97.0} & 0.399 & 0.059 & \textbf{0.025} \\
CFPB complaints (199) & 39.2 & 64.3 & \textbf{86.9} & 0.525 & 0.134 & \textbf{0.043} \\
\quad fraud (80) & 30.0 & 57.5 & \textbf{90.0} & & & \\
\bottomrule
\end{tabular}
\end{table}

The results of \sysplus{} on the three new held-out sets are very encouraging (Table~\ref{tab:plus}). On ABCD, System~X has similar measured accuracy to the base LLM at choosing between acting and replying (51.7\% vs 52.3\%); \sysplus{} reaches 88.7\% with an ECE of 0.037. On the 80 filtered, held-out fraud complaints it matches the mapped action label 90.0\% of the time, against 57.5\% for System~X. Banking77 rises from 53.5\% to 57.2\%.

The broad decision benchmarks do not improve. JevBench accuracy is 52.4\%; against the base LLM, 17 items are won and 24 lost ($p=0.35$), and against System~X 11 are won and 16 lost ($p=0.44$). JevBench ECE is 0.201, worse than System~X's 0.169. This is consistent with LLM-as-Jev's observation that single-domain data can hurt general calibration, here as the reasoning share of the mixture halves. On the hand-written set, \sysplus{} is not better than System~X (76.3\% vs 78.4\%; 2 items won and 4 lost, $p=0.69$). Its fraud cases are still mostly routed to \reply{}: 5 of 10 are correct, and 16 of its 23 errors go to \reply{}, 10 of them at 0.80 confidence or more. Two differences between the new data and this set may explain it, though we have not tested either: CFPB complaints are long narratives written to a regulator, not short live messages, and ABCD's agents verify identity before acting, so asking for details first is often the correct label there.

\section{Reply quality}\label{sec:replyq}

The benchmarks above score decisions. The public decision benchmarks used here do not score the replies, so we use four complementary checks: automatic checks on every reply, ABCD's utterance-ranking task, two LLM judges, and the general generation suite of LLM-as-Jev (\S\ref{sec:general}).

\subsection{Automatic checks}
For every nonempty reply written during the decision evaluations, we run string-based checks. The \emph{fact-copy} check passes when the reply contains every qualifying State fact that also appears in the reference, using case-insensitive substring matching. Qualifying facts are string values of at least four characters in the company, customer or account blocks. The \emph{unsupported identifier} check flags a regex-matched URL, email address, phone number, dollar amount or order-style identifier that is absent from the normalized State. These checks do not establish semantic grounding or measure all hallucinations: unsupported policies, promises and other factual claims can pass them. Numeric formatting can also cause false positives: one v4 reply gives \$3.50 when the State stores 3.5, which the string-based check incorrectly flags. We also count template slots and instruction echoes as in \S\ref{sec:errors}.

\begin{table}[t]
\centering
\caption{Automatic reply checks (\% of replies written). Fact copying is measured only on generated replies whose references contain a qualifying State fact (84, 80 and 84 Bitext-vertical rows for the three models). The evaluated subsets differ.}
\label{tab:checks}
\small
\begin{tabular}{llccc}
\toprule
Set & Check & Base LLM & System~X & \sysplus{} \\
\midrule
Bitext verticals & fact-copy check & 2.4 & 28.8 & \textbf{46.4} \\
 & unsupported identifier & 0.0 & 2.5 & 0.7 \\
 & instruction echo & 66.4 & 17.4 & \textbf{2.6} \\
ABCD & instruction echo & 62.8 & 26.2 & \textbf{0.6} \\
Held-out Bitext & instruction echo & 65.6 & 20.3 & \textbf{16.2} \\
 & template slot & 2.6 & 12.8 & 16.9 \\
\bottomrule
\end{tabular}
\end{table}

Table~\ref{tab:checks} shows that \sysplus{} copies the needed State fact in 46\% of cases, against 29\% for System~X, and almost never echoes the instruction on the new sets. The identifier check flags few outputs; all flagged strings (four outputs for System~X, one for \sysplus{}) are a website of the form \texttt{www.<company>.com}, built from the company name when the State gives none. System~X never saw our synthetic websites, which follow the same pattern, so the appearance of this pattern cannot be attributed solely to the added fine-tuning data. Template slots on the held-out Bitext rows rise to 17\%, while the unchanged System~X reply targets still contain such slots.

\subsection{ABCD utterance ranking}
ABCD's Cascading Dialogue Success task includes utterance ranking: at each agent turn whose next step is a reply, the true agent utterance is hidden among 100 candidates, and systems are scored by Recall@$k$ \citep{chen2021abcd}. We rank the candidates by the model's log-probability of writing each one after \texttt{r] Reply:}, on ABCD's delexicalised text. Raw likelihood strongly favours generic utterances such as ``is there anything else I can help you with?'', so our primary score subtracts the log-probability of the same candidate given an empty conversation, a pointwise mutual information correction for surface-form competition \citep{holtzman2021surface}. The reported evaluation uses 500 randomly sampled test turns that follow the customer's first message. Each turn's option list holds 3--6 randomly sampled ABCD actions plus \reply{}. Our ABCD training rows use the original, non-delexicalised text, so the ranking inputs are slightly out of format for the trained models. For candidate token sequence $z$ (including the end token), the score is
\begin{equation}
s_{\mathrm{PMI}}(z)=\log p_\theta(z\mid x,d_r)-\log p_\theta(z\mid x_0,d_r),
\end{equation}
where $x_0$ keeps the options but replaces conversation and action history with empty lists. The implementation assigns rank $1+\sum_j\mathbb{1}[s_j>s_g]$, so exact ties receive the best rank. Recall@$k$ is the fraction of gold ranks at most $k$, and MRR is their mean reciprocal rank.
We later reran the ranking for the base LLM, System~X and \sysplus{} with the same seed and 500 sampled turns, now saving each turn's gold utterance, its 100 candidate identifiers and its rank under every scoring. The rerun reproduced the aggregate metrics in Table~\ref{tab:rank} exactly, and the per-turn files allow paired comparisons.

\begin{table}[t]
\centering
\caption{ABCD utterance ranking among 100 candidates (500 test turns, PMI scoring, RTX~5090, bf16). The ABCD rows are from \citet{chen2021abcd}, Table~4: retrieval models trained for this task and evaluated on the full test set, averaged over three seeds.}
\label{tab:rank}
\small
\begin{tabular}{lcccc}
\toprule
Model & R@1 & R@5 & R@10 & MRR \\
\midrule
Base LLM & 7.2 & 18.2 & 28.8 & 0.148 \\
System~X & 8.0 & 23.0 & 32.8 & 0.166 \\
\sysplus{} & \textbf{15.0} & \textbf{36.2} & \textbf{49.4} & \textbf{0.263} \\
\midrule
ABCD BERT-base & 21.7 & 46.6 & 58.7 & -- \\
ABCD RoBERTa-Large & 22.0 & 47.8 & 59.1 & -- \\
\bottomrule
\end{tabular}
\end{table}

\sysplus{} doubles the base LLM's Recall@1 (Table~\ref{tab:rank}). Its ABCD training rows come from the same dialogue distribution, so this is an in-domain result. It remains below ABCD's purpose-trained retrieval baselines, which use a different protocol. Without the PMI correction, \sysplus{} reaches only 2.2/12.2/22.0 at $k=1/5/10$ (mean log-probability per token).

\subsection{LLM judges}
We take 150 held-out rows whose gold answer is a reply (50 each from held-out Bitext, Bitext verticals and ABCD) and collect one reply per row from six sources:
\begin{itemize}
\item the base LLM, System~X and \sysplus{}, each forced to reply from the decision cache with a 48-token cap;
\item two \emph{two-call} baselines: a separate Qwen3-0.6B or Qwen3-1.7B call that reads the State again with a plain instruction to write a 1--2 sentence reply, without any options or task adapter, with a 64-token cap;
\item the dataset's reference reply.
\end{itemize}
The judge is the Ollama endpoint \texttt{gemma4:31b-cloud}, identified as Gemma~4 31B \citep{gemma4}, requested at temperature 0 and blind to the source names. Its serving revision is not pinned in the saved results. It scores each reply from 1 to 5 for helpfulness, grounding in the State and clean form, and says whether a support lead would send it; the reference reply is not shown during scoring. In pairwise comparisons it picks the better reply to send; each pair is judged in both orders, and a verdict that changes with the order counts as a tie. A row on which only one of the two orders returned a valid judgment also counts as a tie. Model replies were generated in bf16, on the RTX~5090 for the base LLM and on the M4 Pro for the others.

\begin{table}[t]
\centering
\caption{LLM-judge results on 150 reply rows (Gemma~4 31B). Left: mean scores (1--5) and share the judge would send. Right: \sysplus{}'s pairwise wins, ties (including order flips) and losses against each source, with exact two-sided sign tests. The base-LLM pairwise comparison has 149 usable rows: both judgments failed on one row. All other pairwise comparisons have 150.}
\label{tab:judge}
\small
\setlength{\tabcolsep}{4pt}
\begin{tabular}{lcccc|cc}
\toprule
Source & Helpful & Grounded & Clean & Send (\%) & \sysplus{} W / T / L & Sign test \\
\midrule
Base LLM & 1.44 & 2.93 & 2.53 & 9.3 & 101 / 35 / 13 & $p<0.001$ \\
System~X & 2.27 & 3.83 & 3.89 & 22.7 & 60 / 41 / 49 & $p=0.34$ \\
\sysplus{} & 2.41 & \textbf{4.41} & 4.34 & 20.7 & -- & \\
Two-call Qwen3-0.6B & 1.73 & 4.32 & 2.99 & 9.3 & 98 / 29 / 23 & $p<0.001$ \\
Two-call Qwen3-1.7B & 3.43 & 4.26 & \textbf{4.83} & \textbf{43.3} & 32 / 30 / 88 & $p<0.001$ \\
Reference & \textbf{3.51} & 4.33 & 4.45 & 38.7 & 35 / 21 / 94 & $p<0.001$ \\
\bottomrule
\end{tabular}
\end{table}

Table~\ref{tab:judge} compares the configured reply systems. \sysplus{} wins 98 pairs and loses 23 against the unadapted Qwen3-0.6B reply baseline. The unadapted Qwen3-1.7B baseline wins 88 pairs to 32, with 43.3\% of replies judged sendable versus 20.7\% for \sysplus{}. These comparisons vary fine-tuning, prompt content and generation limits, and therefore do not isolate cache reuse or model size. They are reply-only evaluations on gold-reply rows, rather than end-to-end pipeline evaluations. The matched cache-versus-reprefill latency experiment in Table~\ref{tab:latency} uses identical weights and prompts and remains a separate test. \sysplus{} versus System~X yields $p=0.34$, which does not establish equivalent reply quality. A second rating pass on \sysplus{} gives mean helpfulness 2.43 versus 2.41; this is a limited repeatability check, not evidence of judge accuracy.

A second set of grades, from Claude Opus~5.5 (model ID \texttt{claude-opus-5-5}, run as two fresh agents with no access to the source mapping), covers 100 pairs (25 against each of the base LLM, System~X, the Qwen3-1.7B reply baseline and the reference), spanning 78 distinct rows. Each pair was graded in both orders. The model ID is taken from the agents' transcripts, which also contain two setup-metadata entries naming another model; no grade was generated by it. Inference settings were the agent defaults. The grades give the same verdict after the order swap on 90\% of pairs and agree with Gemma on 69\% (Cohen's $\kappa=0.507$). The agreement is 88.7\% when both judges name a winner. From \sysplus{}'s perspective, the win/loss counts are 18/5 against the base LLM, 12/7 against System~X, 6/16 against the 1.7B baseline and 4/15 against the reference. The 1.7B comparison has $p=0.052$, so this subset does not independently establish a preference at the 5\% level. No human evaluation was performed.

\section{General capabilities}\label{sec:general}

Following LLM-as-Jev, we run the lm-evaluation-harness \citep{biderman2024lmeval} with the paper's sample sizes: GSM8K \citep{cobbe2021gsm8k} (250 items, 5-shot), IFEval \citep{zhou2023ifeval} (200), TriviaQA \citep{joshi2017triviaqa} (300, 5-shot), LAMBADA \citep{paperno2016lambada} (500) and WikiText-2 \citep{merity2017wikitext}. We run IFEval with the chat template and thinking off; the other tasks use the plain prompt format. All runs use bf16 on the RTX~5090.

\begin{table}[t]
\centering
\caption{General capabilities (accuracy in \%, WikiText-2 word perplexity). ``Paper'' is the untrained Qwen3-0.6B row of LLM-as-Jev (its Table~9). GSM8K is strict match; IFEval is prompt-level strict accuracy. Avg.\ is the mean of the four accuracies.}
\label{tab:general}
\small
\begin{tabular}{lccccc|c}
\toprule
Model & GSM8K & IFEval & TriviaQA & LAMBADA & Avg. & WikiText ppl \\
\midrule
Paper: base LLM & 44.4 & 55.0 & 4.0 & 39.0 & 35.6 & 27.52 \\
\midrule
Base LLM & 42.4 & \textbf{55.5} & 12.0 & 39.6 & 37.4 & 26.13 \\
System~X & \textbf{46.4} & 52.0 & 12.7 & \textbf{40.4} & \textbf{37.9} & 24.57 \\
\sysplus{} & 40.0 & 45.0 & \textbf{14.3} & 39.8 & 34.8 & \textbf{24.12} \\
\bottomrule
\end{tabular}
\end{table}

Our base LLM matches the paper's untrained row on GSM8K, IFEval and LAMBADA to within two points (Table~\ref{tab:general}). TriviaQA differs (12.0\% vs 4.0\%); we did not resolve this protocol difference, so we compare our models only with each other. System~X changes by at most four points on each task (we did not test these differences) and lowers perplexity. \sysplus{} loses 10.5 points of IFEval accuracy (55.5\% to 45.0\%; 6 prompts won and 27 lost, exact McNemar $p<0.001$). This resembles the IFEval drop LLM-as-Jev reports for its intent-only 0.6B run (55.0\% to 48.0\%). It suggests that the larger share of short, templated support targets erodes instruction following, despite the reply KL term. GSM8K falls by 2.4 points relative to the base LLM, TriviaQA rises by 2.3 and LAMBADA rises by 0.2. These are descriptive comparisons; we have not performed paired significance tests for them. In particular, being within an estimated error range would not establish equivalence.

\section{A larger backbone and teacher-rewritten replies: \sysfour}\label{sec:v4}

\sysfour{} changes three things at once: the backbone (Qwen3-1.7B instead of Qwen3-0.6B), the data balance, and the reply targets. To separate the backbone from the other two changes, we also train \sysfoursmall, Qwen3-0.6B on the same \sysfour{} data, and evaluate the untrained Qwen3-1.7B as its base LLM.

\subsection{Data and training}
The mixture has 24{,}256 rows, with customer-support rows at 42\% (\sysplus: 66\%):
\begin{itemize}
\item \textbf{Intents:} the same 1{,}000 CLINC150, MASSIVE and Bitext intent rows as System~X.
\item \textbf{Reasoning:} 2{,}000 sampled rows from each of the six reasoning sources, twice the System~X count. Of these 12{,}000 rows, 1{,}303 reuse a State present in the original System~X reasoning mixture; the mixtures are not disjoint.
\item \textbf{Support, initially sampled from the earlier training rows while keeping each source's mix of kinds; the retained counts after rewrite filtering are:}
  \begin{itemize}
  \item Bitext reply task: 1{,}788 rows;
  \item ABCD: 3{,}500;
  \item Bitext verticals: 2{,}968;
  \item CFPB: 1{,}000.
  \end{itemize}
\end{itemize}
Every Bitext and Bitext-vertical reply target was rewritten by a teacher, Qwen3-14B \citep{yang2025qwen3} served by Ollama as \texttt{qwen3:14b} (Q4\_K\_M quantization) with thinking off and temperature 0.
\begin{itemize}
\item \textbf{Teacher input:} the State and the full original Bitext response, including its template slots.
\item \textbf{Teacher instructions:}
  \begin{itemize}
  \item keep the response's intent and main step in one or two sentences;
  \item use State facts, and fill a slot only from the State;
  \item add no URLs, numbers, policies or names of pages or buttons that are not in the State or written in plain text in the reference;
  \item do not ask for information already in the State;
  \item write plain text of at most 40 words, with no greeting, apology, empathy phrase or list.
  \end{itemize}
\item \textbf{Decoding limit:} 100 new tokens per rewrite.
\item \textbf{Post-processing:} we drop empathy and sign-off sentences and keep at most two sentences.
\item \textbf{Checks:} a rewrite is rejected if it is empty, contains an identifier absent from both the State and the full reference, retains a template slot or instruction echo, or exceeds two sentences or 55 words. Its row is then dropped. Unlike the evaluation check in \S\ref{sec:replyq}, the rewrite filter permits identifiers found in the reference as well as the State; it does not ensure that all retained details are grounded in the State. This removed 244 of 2{,}500 rows (9.8\%). Most are rows whose customer message itself contains an unfilled Bitext slot, or whose reference relies on a slot the State cannot fill.
\item \textbf{ABCD replies} are human-written agent messages and were kept as written. In a pilot, the teacher added unsupported details to them, for example a membership extended ``for a month''.
\end{itemize}
Held-out sets and their references are unchanged. Training uses the System~X recipe on one RTX~5090:
\begin{itemize}
\item \textbf{\sysfour:} 758 steps in 57 minutes; 17.4M LoRA parameters;
\item \textbf{\sysfoursmall:} 29 minutes;
\item \textbf{both:} the same seeds as before, a single run each.
\end{itemize}
All \sysfour{}, \sysfoursmall{} and Qwen3-1.7B evaluations ran on the RTX~5090, with the same precision per task as before. For \sysplus{}, decisions measured on the M4 Pro and on the RTX~5090 agreed on 1{,}192 of 1{,}196 reply-task rows; paired tests between a column measured on the M4 Pro and one measured on the RTX~5090 include this small hardware effect.

\subsection{Decisions}

\begin{table}[t]
\centering
\caption{Decisions (argmax accuracy in \%, ECE in parentheses). \sysplus{} and the 0.6B base LLM were measured on the M4 Pro, and the other columns on the RTX~5090. JevBench in FP32; all other sets in bf16.}
\label{tab:v4}
\small
\setlength{\tabcolsep}{3pt}
\begin{tabular}{lccccc}
\toprule
& \multicolumn{3}{c}{Qwen3-0.6B} & \multicolumn{2}{c}{Qwen3-1.7B} \\
\cmidrule(lr){2-4}\cmidrule(lr){5-6}
Set & Base LLM & \sysplus{} & \sysfoursmall{} & Base LLM & \sysfour{} \\
\midrule
JevBench (231) & 55.4 (.292) & 52.4 (.201) & 51.5 (.175) & 55.0 (.404) & \textbf{61.0} (\textbf{.123}) \\
Banking77 (3{,}080) & 18.5 (.460) & 57.2 (.081) & 55.5 (.056) & 28.3 (.617) & \textbf{65.2} (.062) \\
Hand-written (97) & 47.4 (.491) & 76.3 (.139) & 84.5 (.115) & 72.2 (.251) & \textbf{86.6} (\textbf{.070}) \\
Held-out Bitext (300) & 58.7 (.389) & \textbf{99.0} (.019) & 96.3 (.022) & 73.3 (.235) & 97.3 (.025) \\
ABCD next step (300) & 52.3 (.420) & \textbf{88.7} (.037) & 88.3 (.035) & 48.3 (.483) & 86.0 (.048) \\
Bitext verticals (300) & 58.3 (.399) & \textbf{97.0} (.025) & 96.7 (.040) & 74.7 (.226) & \textbf{97.0} (\textbf{.020}) \\
CFPB complaints (199) & 39.2 (.525) & 86.9 (.043) & \textbf{87.9} (.052) & 53.3 (.435) & 85.9 (.049) \\
\quad fraud (80) & 30.0 & 90.0 & \textbf{96.2} & 46.2 & 95.0 \\
\bottomrule
\end{tabular}
\end{table}

\begin{figure}[t]
\centering
\includegraphics[width=\linewidth]{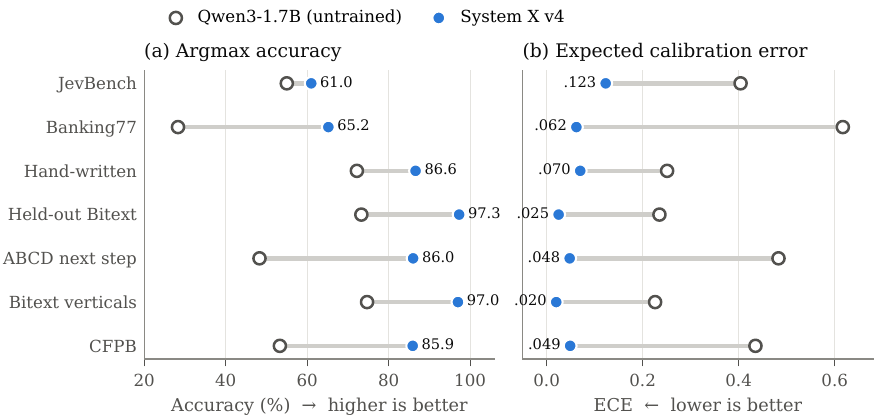}
\caption{\sysfour{} against its untrained Qwen3-1.7B base LLM on every decision set (Table~\ref{tab:v4}; argmax before fallback). Accuracy rises and calibration error falls on all seven sets. JevBench in FP32; all other sets in bf16.}
\label{fig:v4}
\end{figure}

\paragraph{Compared with its own base LLM (Table~\ref{tab:v4}),} \sysfour{} yielded very encouraging results across the board:
\begin{itemize}
\item JevBench accuracy: 61.0\% against 55.0\% (27 items won, 13 lost, $p=0.04$), with ECE 0.123 against 0.404;
\item Banking77: 65.2\% against 28.3\%;
\item the hand-written set: 86.6\% against 72.2\% (19 won, 5 lost, $p=0.007$);
\item ABCD: 86.0\% against 48.3\% (127 won, 14 lost).
\end{itemize}
Its JevBench accuracy equals the best 0.6B result reported by LLM-as-Jev, though on a larger backbone and under different training conditions. Figure~\ref{fig:v4} shows these gains, together with the calibration changes, on every decision set.

\paragraph{The 0.6B control evaluates the combined data changes.} \sysfoursmall{} keeps the support-task results of \sysplus{}: ABCD 88.3\% against 88.7\% (5 won, 6 lost), and CFPB 87.9\% against 86.9\%. It is better on the hand-written set: 84.5\% against 76.3\% (10 won, 2 lost, $p=0.04$). JevBench does not recover (51.5\%; against the 0.6B base LLM, 17 won and 26 lost, $p=0.22$). The JevBench gain is observed only with the larger backbone in these runs. The hand-written gain appears at both sizes, so it is associated with the revised training recipe; data balance and teacher rewriting remain confounded.

\paragraph{Errors.} On the hand-written set, \sysfour{} gets:
\begin{itemize}
\item \textbf{actions it previously routed to \reply{}:} 9 of 10 fraud cases, 9 of 9 escalations and 10 of 10 refunds (System~X: 6, 3 and 5);
\item \textbf{state conflicts:} still only 2 of 8, its main remaining failure. It acts on, for example, an already-shipped order or an inactive subscription.
\end{itemize}
Against System~X on this set it wins 15 cases and loses 7 ($p=0.13$).

\subsection{Replies}

\paragraph{Automatic checks.}
\begin{itemize}
\item \textbf{Fact copying (Bitext verticals):} 70.4\% for \sysfour{} (81 rows), against 57.8\% for \sysfoursmall{} (83), 46.4\% for \sysplus{} and 28.0\% for its base LLM (75).
\item \textbf{Instruction echo:} \sysfour{} produces none on any set.
\item \textbf{Template slots:} all 14 in its held-out Bitext replies are copied from slots already present in those customer messages.
\item \textbf{Unsupported identifiers:} the check flags 2\% of its held-out Bitext replies. These are invented phone numbers of the placeholder form \texttt{1-800-123-4567}, written when a customer asks for a phone number that the State does not contain. The same failure appears in the two other models evaluated here.
\end{itemize}

\paragraph{ABCD utterance ranking.} Under the protocol of Table~\ref{tab:rank}:

\begin{center}
\small
\begin{tabular}{lcccc}
\toprule
Model & R@1 & R@5 & R@10 & MRR \\
\midrule
Qwen3-1.7B base LLM & 2.2 & 8.0 & 15.8 & 0.075 \\
\sysfoursmall{} & 13.2 & 37.2 & 50.6 & 0.255 \\
\sysfour{} & \textbf{17.6} & \textbf{39.6} & \textbf{53.8} & \textbf{0.288} \\
\bottomrule
\end{tabular}
\end{center}

\sysfour{} reaches 17.6/39.6/53.8, the closest of our models to ABCD's purpose-trained retrievers (21.7--22.0 at $k=1$). The untrained Qwen3-1.7B ranks worse than the untrained 0.6B model under this scoring.

\paragraph{LLM judges.} We use the same 150 rows and Gemma protocol as Table~\ref{tab:judge}, with \sysfour{} as the focus system. Replies from the untrained Qwen3-1.7B, forced through the decision cache, were generated on the M4 Pro; those from \sysfour{} and \sysfoursmall{} on the RTX~5090.

\begin{table}[t]
\centering
\caption{Gemma judge on 150 reply rows with \sysfour{} as the focus system: mean scores (1--5), share judged sendable, and \sysfour{}'s pairwise wins / ties (including order flips) / losses with exact two-sided sign tests. The 1.7B two-call baseline is the same unadapted Qwen3-1.7B reply call as in Table~\ref{tab:judge}.}
\label{tab:judge4}
\small
\setlength{\tabcolsep}{4pt}
\begin{tabular}{lcccc|cc}
\toprule
Source & Helpful & Grounded & Clean & Send (\%) & v4 W / T / L & Sign test \\
\midrule
\sysplus{} & 2.40 & \textbf{4.40} & 4.33 & 20.0 & 99 / 26 / 25 & $p<0.001$ \\
\sysfoursmall{} & 2.93 & 4.35 & 4.65 & 38.0 & 78 / 39 / 33 & $p<0.001$ \\
1.7B base (cache) & 2.96 & 4.33 & 4.27 & 40.0 & 84 / 33 / 33 & $p<0.001$ \\
Two-call Qwen3-1.7B & 3.45 & 4.26 & 4.83 & 44.0 & 66 / 36 / 48 & $p=0.11$ \\
Reference & 3.49 & 4.31 & 4.43 & 36.7 & 79 / 27 / 44 & $p=0.002$ \\
\sysfour{} & \textbf{3.92} & 4.33 & \textbf{4.86} & \textbf{56.7} & -- & \\
\bottomrule
\end{tabular}
\end{table}

\begin{figure}[t]
\centering
\includegraphics[width=\linewidth]{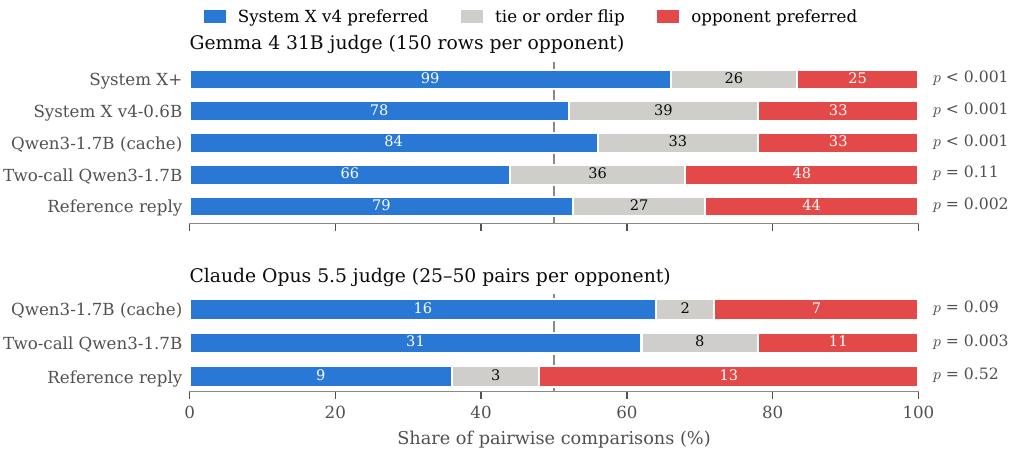}
\caption{Pairwise reply comparisons with \sysfour{} as the focus system: shares of pairs it wins, ties (including order flips) and loses, with counts inside the bars and exact two-sided sign tests on the right. Top: Gemma on 150 rows per opponent (Table~\ref{tab:judge4}). Bottom: Claude on its blind subset of 25 or 50 pairs per opponent. The vertical line marks 50\%.}
\label{fig:judges}
\end{figure}

Table~\ref{tab:judge4} reports Gemma's view:
\begin{itemize}
\item \sysfour{} has the highest helpfulness and sendable share of any source we evaluated;
\item it is preferred to the dataset references (79 to 44, $p=0.002$) and to \sysplus{} (99 to 25);
\item against the unadapted Qwen3-1.7B second call, Gemma's preference is not significant (66 to 48, $p=0.11$).
\end{itemize}
Figure~\ref{fig:judges} shows the pairwise outcomes of both judges.

Claude Opus~5.5 (model ID \texttt{claude-opus-5-5}) graded a second, blind set of 100 pairs in both orders: 50 against the two-call Qwen3-1.7B, 25 against the untrained Qwen3-1.7B and 25 against the reference.
\begin{itemize}
\item \textbf{Against the two-call Qwen3-1.7B:} it prefers \sysfour{}, 31 pairs to 11 ($p=0.003$).
\item \textbf{Against the untrained Qwen3-1.7B:} 16 to 7 ($p=0.09$).
\item \textbf{Against the reference:} 9 to 13 ($p=0.52$).
\item \textbf{Consistency:} its verdicts survive the order swap on 89\% of pairs.
\item \textbf{Agreement with Gemma:} 68\% overall ($\kappa=0.44$), and 85.5\% where both name a winner.
\end{itemize}
These comparisons differ in fine-tuning, prompt and generation limits, as in \S\ref{sec:replyq}. One judge prefers \sysfour{}'s replies to the separate unadapted 1.7B call (Claude, 31 to 11, $p=0.003$); the other's preference is not significant (Gemma, 66 to 48, $p=0.11$). Neither establishes equivalence, and \sysfour{} also saves the second prefill. They do not show that cache reuse improves quality: for fixed weights and prompt, the cached reply equals a re-read reply, and in FP32 we again observe identical tokens on the demonstration cases. The reply-quality gains accompany a changed backbone, data balance and teacher-rewritten targets. The teacher supplies supervision from a larger model, but its contribution has not been isolated by a matched run without rewriting.

\paragraph{Latency at 1.7B.} With the protocol of Table~\ref{tab:latency} on the M4 Pro, timed with the untrained Qwen3-1.7B weights (merging LoRA preserves the inference architecture, but the adapted weights were not timed in this test), replying from the cache saves 178\,ms (17\%) at 367 prompt tokens, 570\,ms (29\%) at 1{,}159 and 2{,}279\,ms (39\%) at 4{,}161 tokens. That is slightly more, relative to end-to-end time, than at 0.6B.

\begin{figure}[t]
\centering
\includegraphics[width=0.62\linewidth]{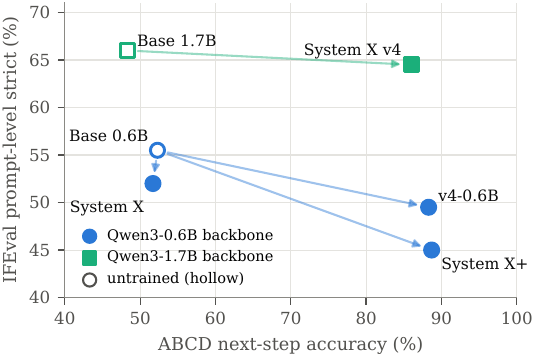}
\caption{Support-task gains against instruction following. Each model is placed by its held-out ABCD next-step accuracy (Tables~\ref{tab:plus} and~\ref{tab:v4}) and IFEval prompt-level strict accuracy (Table~\ref{tab:general} and \S\ref{sec:v4}); arrows run from each untrained base LLM to the models fine-tuned from it. At 0.6B, the support gains of \sysplus{} and \sysfoursmall{} come with lower IFEval scores. At 1.7B, \sysfour{} reaches similar ABCD accuracy with an IFEval change that is not statistically significant.}
\label{fig:tradeoff}
\end{figure}

\paragraph{General capabilities.} Under the protocol of Table~\ref{tab:general}, \sysfour{} stays close to its base LLM:
\begin{itemize}
\item GSM8K 69.2\% against 68.0\%;
\item IFEval 64.5\% against 66.0\% (13 prompts won, 16 lost, $p=0.71$);
\item TriviaQA 27.7\% against 27.3\%;
\item LAMBADA 48.6\% against 50.0\%;
\item WikiText perplexity 16.7 against 21.1.
\end{itemize}
The measured IFEval decrease at 1.7B is 1.5 percentage points and is not statistically significant ($p=0.71$). This does not establish that instruction-following ability is preserved. At 0.6B, \sysfoursmall{} scores 49.5\% on IFEval: numerically above \sysplus{}'s 45.0\% (18 won, 9 lost, $p=0.12$), but still below the 0.6B base LLM's 55.5\% (7 won, 19 lost, $p=0.03$). Its other scores are GSM8K 40.0\%, TriviaQA 13.3\%, LAMBADA 38.8\% and perplexity 23.7. Figure~\ref{fig:tradeoff} places each model by its ABCD next-step accuracy and IFEval score.

\section{Limitations}
\begin{itemize}
\item \textbf{Scale and variance.} We use two model sizes (0.6B and 1.7B), LoRA only and a single seed per configuration; seed variance is unmeasured. \sysfour{} changes backbone, data balance and reply targets together, and \sysfoursmall{} separates only the backbone. The 231 public JevBench items provide limited power to resolve differences of a few points.
\item \textbf{Deviations from LLM-as-Jev.} We do not anchor off-path trie nodes or reproduce the source paper's cross-benchmark n-gram decontamination procedure. Our explicit message-overlap exclusion applies to the Bitext split. We do not include the paper's long-context sets (ContractNLI \citep{koreeda2021contractnli}, MuSiQue \citep{trivedi2022musique}, QuALITY \citep{pang2022quality}). We train with LoRA where the paper's Reason-9k row uses full fine-tuning. We ran Banking77 in bf16.
\item \textbf{The hand-written set.} It is small (97 cases), and its authors also wrote the template data for an earlier version of this model. It measures the failure modes above more than it estimates deployment accuracy. The v3 and v4 recipes were developed after inspecting earlier results on this set; it is therefore a repeatedly consulted diagnostic set, rather than an independent final test.
\item \textbf{Reply data.} Bitext is hybrid synthetic, and a message-level split does not test generalisation to independent data sources or template families. The original System~X reply rows have no account state; \sysplus{} adds State facts but still lacks a controlled evaluation of semantic grounding. The responses are templated, and ROUGE-L against them is a weak proxy for usefulness. Reply quality is assessed by two LLM judges that agree only moderately ($\kappa=0.44$--$0.51$); we have no human evaluation of replies.
\item \textbf{\sysplus{} data.} The added sources are labelled by our own mappings: ABCD actions to renamed options, CFPB complaint categories to actions (filtered by keyword cues), and Bitext intents to actions. The vertical States use synthetic facts, 113 targets contain a slot-filling error, and ABCD is in-domain for the utterance-ranking result. \sysplus{} trained for twice as many steps as System~X, so data and schedule are confounded, and v4 changes support share together with reply targets, so it does not isolate the cause of the IFEval change. A stronger reply KL has not been tested separately.
\item \textbf{Missing component ablations.} We lack a decision-only LoRA run matched for data, compute and seed, so the current results cannot isolate the effect of reply training. We have not separated the effects of the Brier term, the reply KL or the reply sentence in the prompt, and we have not compared against a reranker baseline such as Qwen3-Reranker-0.6B. Our two-call baselines measure reply quality only; we did not time a pipeline with a larger second model.
\item \textbf{Data licenses.} ReClor restricts use to non-commercial research and LogiQA~2.0 uses CC~BY-NC-SA~4.0, and the adapter remains private. Release terms and compatibility across the training datasets still need review.
\end{itemize}

\section{Conclusion}
In the present study, System~X combines a decision model and text generation in one LLM. The \reply{} option selects a text response, which continues from the KV cache already built for the decision. This avoids a repeated prompt prefill. Joint training improves held-out Bitext routing and calibration, while the measured JevBench accuracy difference is not statistically significant. A matched decision-only training baseline is needed to determine the cost of adding reply training. System~X still routes unfamiliar actions to replies and can select actions that contradict account state.

The run with added crowdworker dialogue, industry and complaint data (\sysplus{}) raises held-out ABCD next-step accuracy from 51.7\% to 88.7\% and fraud-complaint accuracy from 57.5\% to 90.0\%, and makes replies copy State facts more often. Its ABCD utterance-ranking Recall@1 rises from 8.0\% to 15.0\% on the 500-turn sample. It does not improve JevBench or the hand-written set, JevBench calibration worsens, and IFEval accuracy falls by 10.5 points, so this training configuration trades improved support-task results for worse measured instruction following. On reply quality, Gemma prefers the trained 0.6B reply system to the unadapted 0.6B reply baseline and prefers the unadapted 1.7B baseline to both. These comparisons change prompts and training as well as model size. Cache reuse saves latency under the matched protocol; its effect on quality is not isolated by the judge experiment. Our next steps are replies grounded in account state and policy, larger backbones, the long-context sets, ablations, and a human evaluation of replies.

\sysfour{} (\S\ref{sec:v4}) moves to Qwen3-1.7B, rebalances the mixture and trains on teacher-rewritten reply targets. Against its base LLM, it improves JevBench, Banking77 and the hand-written set, shows no statistically significant IFEval change, and is accompanied by a latency test on the unadapted 1.7B backbone. Against an unadapted Qwen3-1.7B second call, one LLM judge prefers its replies significantly (Claude, $p=0.003$) and the other's preference is not significant (Gemma, $p=0.11$). Its 0.6B control keeps the support-task gains and improves the hand-written set, but not JevBench. State conflicts remain the main decision failure, and invented placeholder phone numbers the main reply failure. Seed variance, a human evaluation of replies and a controlled ablation of the teacher rewrite are still missing. Further studies with several seeds, human raters and real customer-support data are recommended. The findings would further pave the pathway for small, single-model agents that decide and reply as an adjunct tool for customer-support teams.

\paragraph{Code, models and reproducibility.} The code is publicly available at \url{https://github.com/Turinglogicuk/system-x}. It contains the data preparation, training, evaluation, latency and judging scripts, a dependency lockfile and run instructions; the per-example evaluation outputs can be regenerated with these scripts. The figures are drawn from the reported tables by \texttt{figures/make\_figures.py} in the paper source. The trained \sysfour{} LoRA adapter is released on Hugging Face at \url{https://huggingface.co/TuringLogic/system-x} under the proposed CC~BY-NC~4.0 terms (non-commercial use only), subject to resolving the training-data licence questions in Appendix~\ref{app:licenses}. These proposed adapter terms do not by themselves establish compatibility with every upstream dataset licence. The base models are not redistributed.

\paragraph{Acknowledgements.} We build directly on the LLM-as-Jev prompt, readout, objective and data recipe of \citet{li2026llmasjev}. The base models Qwen3-0.6B and Qwen3-1.7B and the teacher Qwen3-14B are by the Qwen team (Apache-2.0). Jev and JevBench inspired this work; it is not affiliated with TypeSafe AI.

\bibliographystyle{plainnat}
\bibliography{refs}

@misc{almeida2026jev,
  author       = {Diogo Almeida},
  title        = {Introducing System One Models \& {Jev}},
  howpublished = {TypeSafe AI blog, \url{https://typesafe.ai/blog/introducing-system-one-models-and-jev}},
  year         = {2026},
  note = {Accessed 15 October 2026}
}

@misc{jevbench2026,
  author       = {{Benchmark Heaven}},
  title        = {{JevBench}: Benchmark for AI Decision Models},
  howpublished = {Public benchmark repository, \url{https://github.com/fstandhartinger/jevbench}},
  year         = {2026},
  note = {Accessed 15 October 2026}
}

@article{li2026llmasjev,
  author  = {Yinheng Li and Justin Wagle},
  title   = {{LLM-as-Jev}: {LLMs} Are Already {Jev}-Style Decision Models---When and How to Fine-Tune Them},
  journal = {arXiv preprint arXiv:2610.02076},
  year    = {2026},
  url = {https://arxiv.org/abs/2610.02076v3}
}

@article{zhang2026anyjev,
  author  = {Jiamu Zhang and Tianze Yang and Yucheng Shi and Evan Chen and Zixiang Nie and Kelly Wan and Liangjie Hong and Ninghao Liu and Liang Wu},
  title   = {{AnyJev} Technical Report},
  journal = {arXiv preprint arXiv:2610.00831},
  year    = {2026},
  url = {https://arxiv.org/abs/2610.00831}
}

@misc{anyjevrepo,
  author       = {{Nokia Applied Research}},
  title        = {{AnyJev} code release (\texttt{anyjev/tacit.py}, \texttt{adaptive=True} mode)},
  howpublished = {\url{https://github.com/nokia-applied-research/AnyJev/blob/main/anyjev/tacit.py}},
  year         = {2026},
  note = {Accessed 15 October 2026}
}

@misc{kwok2026clm,
  author       = {Jacky Kwok and Hangoo Kang and Tarun Suresh and Jon Saad-Falcon and Marco Pavone and Christopher R{\'e} and Azalia Mirhoseini},
  title        = {Contrastive Language Models},
  howpublished = {Notion blog; author-provided citation and code at \url{https://github.com/Contrastive-LM/CLM}},
  year         = {2026},
  note = {Accessed 15 October 2026}
}

@article{li2026jevjudge,
  author  = {Yubo Li and Yidi Miao and Ramayya Krishnan and Rema Padman},
  title   = {{JEV}-as-a-Judge: Accept When Confident, Escalate When Unsure},
  journal = {arXiv preprint arXiv:2609.26550},
  year    = {2026},
  url = {https://arxiv.org/abs/2609.26550v5}
}

@article{rao2026rubric,
  author  = {Delip Rao and Chris Callison-Burch},
  title   = {Decision Models vs.\ {LLMs} as Rubric Judges: Cheaper, Faster, and Wrong in the Same Places},
  journal = {arXiv preprint arXiv:2609.29769},
  year    = {2026},
  url = {https://arxiv.org/abs/2609.29769v3}
}

@misc{lee2026semif,
  author       = {Theodore Lee},
  title        = {{SemIf} (formerly {OpenJev})},
  howpublished = {\url{https://github.com/TheoLeeCJ/SemIf-OpenJev}},
  year         = {2026},
  note = {Accessed 15 October 2026}
}

@article{yang2025qwen3,
  author  = {An Yang and others},
  title   = {{Qwen3} Technical Report},
  journal = {arXiv preprint arXiv:2505.09388},
  year    = {2025},
  url = {https://arxiv.org/abs/2505.09388}
}

@inproceedings{hu2022lora,
  author    = {Edward J. Hu and Yelong Shen and Phillip Wallis and Zeyuan Allen-Zhu and Yuanzhi Li and Shean Wang and Lu Wang and Weizhu Chen},
  title     = {{LoRA}: Low-Rank Adaptation of Large Language Models},
  booktitle = {International Conference on Learning Representations},
  year      = {2022},
  url = {https://openreview.net/forum?id=nZeVKeeFYf9}
}

@inproceedings{guo2017calibration,
  author    = {Chuan Guo and Geoff Pleiss and Yu Sun and Kilian Q. Weinberger},
  title     = {On Calibration of Modern Neural Networks},
  booktitle = {Proceedings of the 34th International Conference on Machine Learning},
  pages     = {1321--1330},
  year      = {2017},
  url = {https://proceedings.mlr.press/v70/guo17a.html}
}

@article{brier1950,
  author  = {Glenn W. Brier},
  title   = {Verification of Forecasts Expressed in Terms of Probability},
  journal = {Monthly Weather Review},
  volume  = {78},
  number  = {1},
  pages   = {1--3},
  year    = {1950},
  url = {https://journals.ametsoc.org/view/journals/mwre/78/1/1520-0493_1950_078_0001_vofeit_2_0_co_2.xml}
}

@inproceedings{naeini2015ece,
  author    = {Mahdi Pakdaman Naeini and Gregory F. Cooper and Milos Hauskrecht},
  title     = {Obtaining Well Calibrated Probabilities Using {Bayesian} Binning},
  booktitle = {Proceedings of the AAAI Conference on Artificial Intelligence},
  year      = {2015},
  url = {https://ojs.aaai.org/index.php/AAAI/article/view/9602}
}

@inproceedings{larson2019clinc,
  author    = {Stefan Larson and Anish Mahendran and Joseph J. Peper and Christopher Clarke and Andrew Lee and Parker Hill and Jonathan K. Kummerfeld and Kevin Leach and Michael A. Laurenzano and Lingjia Tang and Jason Mars},
  title     = {An Evaluation Dataset for Intent Classification and Out-of-Scope Prediction},
  booktitle = {Proceedings of EMNLP-IJCNLP},
  pages     = {1311--1316},
  year      = {2019},
  url = {https://aclanthology.org/D19-1131/}
}

@inproceedings{fitzgerald2023massive,
  author    = {Jack FitzGerald and Christopher Hench and Charith Peris and Scott Mackie and Kay Rottmann and Ana Sanchez and Aaron Nash and Liam Urbach and Vishesh Kakarala and Richa Singh and Swetha Ranganath and Laurie Crist and Misha Britan and Wouter Leeuwis and Gokhan Tur and Prem Natarajan},
  title     = {{MASSIVE}: A 1{M}-Example Multilingual Natural Language Understanding Dataset with 51 Typologically-Diverse Languages},
  booktitle = {Proceedings of the 61st Annual Meeting of the Association for Computational Linguistics},
  pages     = {4277--4302},
  year      = {2023},
  url = {https://aclanthology.org/2023.acl-long.235/}
}

@misc{bitext2024,
  author       = {Bitext},
  title        = {Bitext -- Customer Service Tagged Training Dataset for {LLM}-based Virtual Assistants},
  howpublished = {\url{https://huggingface.co/datasets/bitext/Bitext-customer-support-llm-chatbot-training-dataset}},
  year         = {2024},
  note = {Accessed 15 October 2026}
}

@inproceedings{casanueva2020banking77,
  author    = {I{\~n}igo Casanueva and Tadas Tem{\v{c}}inas and Daniela Gerz and Matthew Henderson and Ivan Vuli{\'c}},
  title     = {Efficient Intent Detection with Dual Sentence Encoders},
  booktitle = {Proceedings of the 2nd Workshop on Natural Language Processing for Conversational AI},
  pages = {38--45},
  year      = {2020},
  url = {https://aclanthology.org/2020.nlp4convai-1.5/}
}

@inproceedings{yu2020reclor,
  author    = {Weihao Yu and Zihang Jiang and Yanfei Dong and Jiashi Feng},
  title     = {{ReClor}: A Reading Comprehension Dataset Requiring Logical Reasoning},
  booktitle = {International Conference on Learning Representations},
  year      = {2020},
  url = {https://openreview.net/forum?id=HJgJtT4tvB}
}

@article{liu2023logiqa2,
  author  = {Hanmeng Liu and Jian Liu and Leyang Cui and Zhiyang Teng and Nan Duan and Ming Zhou and Yue Zhang},
  title   = {{LogiQA} 2.0---An Improved Dataset for Logical Reasoning in Natural Language Understanding},
  journal = {IEEE/ACM Transactions on Audio, Speech, and Language Processing},
  volume  = {31},
  pages   = {2947--2962},
  year    = {2023},
  url = {https://doi.org/10.1109/TASLP.2023.3293046}
}

@inproceedings{huang2019cosmosqa,
  author    = {Lifu Huang and Ronan Le Bras and Chandra Bhagavatula and Yejin Choi},
  title     = {Cosmos {QA}: Machine Reading Comprehension with Contextual Commonsense Reasoning},
  booktitle = {Proceedings of EMNLP-IJCNLP},
  pages     = {2391--2401},
  year      = {2019},
  url = {https://aclanthology.org/D19-1243/}
}

@inproceedings{bisk2020piqa,
  author    = {Yonatan Bisk and Rowan Zellers and Ronan Le Bras and Jianfeng Gao and Yejin Choi},
  title     = {{PIQA}: Reasoning about Physical Commonsense in Natural Language},
  booktitle = {Proceedings of the AAAI Conference on Artificial Intelligence},
  volume    = {34},
  pages     = {7432--7439},
  year      = {2020},
  url = {https://ojs.aaai.org/index.php/AAAI/article/view/6239}
}

@inproceedings{bhagavatula2020anli,
  author    = {Chandra Bhagavatula and Ronan Le Bras and Chaitanya Malaviya and Keisuke Sakaguchi and Ari Holtzman and Hannah Rashkin and Doug Downey and Scott Wen-tau Yih and Yejin Choi},
  title     = {Abductive Commonsense Reasoning},
  booktitle = {International Conference on Learning Representations},
  year      = {2020},
  url = {https://openreview.net/pdf/48fb3337a97a1f5f932d55dcebbbadf13ff197e8.pdf}
}

@inproceedings{tafjord2021proofwriter,
  author    = {Oyvind Tafjord and Bhavana Dalvi and Peter Clark},
  title     = {{ProofWriter}: Generating Implications, Proofs, and Abductive Statements over Natural Language},
  booktitle = {Findings of the Association for Computational Linguistics: ACL-IJCNLP 2021},
  pages     = {3621--3634},
  year      = {2021},
  url = {https://aclanthology.org/2021.findings-acl.317/}
}

@inproceedings{koreeda2021contractnli,
  author    = {Yuta Koreeda and Christopher Manning},
  title     = {{ContractNLI}: A Dataset for Document-level Natural Language Inference for Contracts},
  booktitle = {Findings of the Association for Computational Linguistics: EMNLP 2021},
  pages     = {1907--1919},
  year      = {2021},
  url = {https://aclanthology.org/2021.findings-emnlp.164/}
}

@article{trivedi2022musique,
  author  = {Harsh Trivedi and Niranjan Balasubramanian and Tushar Khot and Ashish Sabharwal},
  title   = {{MuSiQue}: Multihop Questions via Single-hop Question Composition},
  journal = {Transactions of the Association for Computational Linguistics},
  volume  = {10},
  pages   = {539--554},
  year    = {2022},
  url = {https://aclanthology.org/2022.tacl-1.31/}
}

@inproceedings{pang2022quality,
  author    = {Richard Yuanzhe Pang and Alicia Parrish and Nitish Joshi and Nikita Nangia and Jason Phang and Angelica Chen and Vishakh Padmakumar and Johnny Ma and Jana Thompson and He He and Samuel R. Bowman},
  title     = {{QuALITY}: Question Answering with Long Input Texts, Yes!},
  booktitle = {Proceedings of NAACL-HLT},
  pages     = {5336--5358},
  year      = {2022},
  url = {https://aclanthology.org/2022.naacl-main.391/}
}

@inproceedings{lin2004rouge,
  author    = {Chin-Yew Lin},
  title     = {{ROUGE}: A Package for Automatic Evaluation of Summaries},
  booktitle = {Text Summarization Branches Out},
  pages     = {74--81},
  year      = {2004},
  url = {https://aclanthology.org/W04-1013/}
}

@inproceedings{chen2021abcd,
  author    = {Derek Chen and Howard Chen and Yi Yang and Alexander Lin and Zhou Yu},
  title     = {Action-Based Conversations Dataset: A Corpus for Building More In-Depth Task-Oriented Dialogue Systems},
  booktitle = {Proceedings of the 2021 Conference of the North American Chapter of the Association for Computational Linguistics: Human Language Technologies},
  year      = {2021},
  pages     = {3002--3017},
  url       = {https://aclanthology.org/2021.naacl-main.239/}
}

@misc{bitextverticals2024,
  author       = {Bitext},
  title        = {Bitext vertical customer-service datasets for {LLM}-based virtual assistants (retail banking, insurance, telco, retail e-commerce, travel, mortgage and loans, wealth management, hospitality, events and ticketing, media)},
  howpublished = {\url{https://huggingface.co/bitext}},
  year         = {2024}
}

@misc{cfpb2026,
  author       = {{Consumer Financial Protection Bureau}},
  title        = {Consumer Complaint Database},
  howpublished = {\url{https://www.consumerfinance.gov/data-research/consumer-complaints/}; parquet mirror \url{https://huggingface.co/datasets/BEE-spoke-data/consumer-finance-complaints}},
  year         = {2026}
}

@inproceedings{holtzman2021surface,
  author    = {Ari Holtzman and Peter West and Vered Shwartz and Yejin Choi and Luke Zettlemoyer},
  title     = {Surface Form Competition: Why the Highest Probability Answer Isn't Always Right},
  booktitle = {Proceedings of the 2021 Conference on Empirical Methods in Natural Language Processing},
  year      = {2021},
  url = {https://aclanthology.org/2021.emnlp-main.564/}
}

@article{zhou2023ifeval,
  author  = {Jeffrey Zhou and Tianjian Lu and Swaroop Mishra and Siddhartha Brahma and Sujoy Basu and Yi Luan and Denny Zhou and Le Hou},
  title   = {Instruction-Following Evaluation for Large Language Models},
  journal = {arXiv preprint arXiv:2311.07911},
  year    = {2023},
  url = {https://arxiv.org/abs/2311.07911}
}

@article{cobbe2021gsm8k,
  author  = {Karl Cobbe and Vineet Kosaraju and Mohammad Bavarian and Mark Chen and Heewoo Jun and Lukasz Kaiser and Matthias Plappert and Jerry Tworek and Jacob Hilton and Reiichiro Nakano and Christopher Hesse and John Schulman},
  title   = {Training Verifiers to Solve Math Word Problems},
  journal = {arXiv preprint arXiv:2110.14168},
  year    = {2021},
  url = {https://arxiv.org/abs/2110.14168}
}

@inproceedings{joshi2017triviaqa,
  author    = {Mandar Joshi and Eunsol Choi and Daniel S. Weld and Luke Zettlemoyer},
  title     = {{TriviaQA}: A Large Scale Distantly Supervised Challenge Dataset for Reading Comprehension},
  booktitle = {Proceedings of the 55th Annual Meeting of the Association for Computational Linguistics},
  year      = {2017},
  url = {https://aclanthology.org/P17-1147/}
}

@inproceedings{paperno2016lambada,
  author    = {Denis Paperno and Germ{\'a}n Kruszewski and Angeliki Lazaridou and Quan Ngoc Pham and Raffaella Bernardi and Sandro Pezzelle and Marco Baroni and Gemma Boleda and Raquel Fern{\'a}ndez},
  title     = {The {LAMBADA} Dataset: Word Prediction Requiring a Broad Discourse Context},
  booktitle = {Proceedings of the 54th Annual Meeting of the Association for Computational Linguistics},
  year      = {2016},
  url = {https://aclanthology.org/P16-1144/}
}

@inproceedings{merity2017wikitext,
  author    = {Stephen Merity and Caiming Xiong and James Bradbury and Richard Socher},
  title     = {Pointer Sentinel Mixture Models},
  booktitle = {International Conference on Learning Representations},
  year      = {2017},
  url = {https://arxiv.org/abs/1609.07843}
}

@article{biderman2024lmeval,
  author  = {Stella Biderman and others},
  title   = {Lessons from the Trenches on Reproducible Evaluation of Language Models},
  journal = {arXiv preprint arXiv:2405.14782},
  year    = {2024},
  url = {https://arxiv.org/abs/2405.14782}
}

@article{gemma4,
  author  = {{Gemma Team}},
  title   = {Gemma 4 Technical Report},
  journal = {arXiv preprint arXiv:2607.02770},
  year    = {2026},
  url     = {https://arxiv.org/abs/2607.02770},
  note    = {Judge accessed as \texttt{gemma4:31b-cloud} through Ollama}
}

\appendix
\section{Prompt example}\label{app:prompt}
An inference prompt as rendered by the implementation (\texttt{build\_prompt}) with the Qwen3 chat template, thinking off and no system message. The options and descriptions are from the engine's catalogue; line breaks marked $\hookleftarrow$ were added for readability, and all other characters are exact. This example includes account state; the Bitext training rows contain only the customer message:
\begin{small}
\begin{Verbatim}[commandchars=@^~]
<|im_start|>user
State: {"customer_msg": "I want to change the delivery address on order 4417.",@cont^~
"order_status": "processing"}
Options:
[1] UPDATE_ADDRESS: Change the shipping address on an order that has not shipped
[2] CANCEL_SUBSCRIPTION: Cancel the customer's subscription
[3] TRACK_ORDER: Send the live tracking link for an order
[4] REPLY_WITH_TEXT: Write a 1-2 sentence reply (answer a question, or ask for@cont^~
missing details)
Select one option. Answer only with its bracketed numeric identifier. If you@cont^~
select REPLY_WITH_TEXT, write a 1-2 sentence reply to the customer after the@cont^~
identifier.<|im_end|>
<|im_start|>assistant
<think>

</think>

Best answer: [
\end{Verbatim}
\end{small}
The assistant turn is pre-filled with \texttt{Best answer: [}. When the reply option is chosen, the continuation is \texttt{4] Reply:} followed by the reply text and the end-of-turn token.

\newpage
\section{Example replies}\label{app:examples}
Held-out Bitext rows whose gold label is \reply{}, with greedy System~X output (bf16) and Bitext's reference:
\begin{small}
\begin{itemize}
\item \emph{``want help to report issues with signup''}\\
\textbf{System~X:} To assist you, could you please provide more details about the issue you're facing?\\
\textbf{Bitext:} To assist you better, please provide us with more details about the specific issues you encountered. \dots
\item \emph{``help me see the early termination penalties''}\\
\textbf{System~X:} To assist you better, could you please provide more details about the specific termination you're referring to?\\
\textbf{Bitext:} When a contract is terminated before the agreed-upon end date, there may be fees or charges involved. \dots
\item \emph{``i dont know what to do tyo see the withdrawal fees''} (instruction echo)\\
\textbf{System~X:} Write a 1-2 sentence reply to the customer.\\
\textbf{Bitext:} To view the withdrawal fees, you can visit our website and navigate to the ``Fees and Charges'' section. \dots
\end{itemize}
\end{small}
Hand-written error, routed to \reply{} at 0.96 instead of \texttt{BLOCK\_FRAUD}:
\begin{small}
\begin{itemize}
\item \emph{``Recieved shipping confirmation for 3 laptops?? I didnt order any laptops''} (\texttt{risk\_score} 0.9)\\
\textbf{System~X:} I'm not sure about the laptops you mentioned. Could you please provide more details about the order?
\end{itemize}
\end{small}
\sysfour{} on the three demonstration cases (FP32, $\tau=0.65$, options \texttt{AUTO\_REFUND}, \texttt{BLOCK\_FRAUD}, \texttt{ESCALATE\_HUMAN}, \reply{}). In the two cases that generate text, the cached reply is token-for-token identical to a full re-read; the first case returns an action without a reply:
\begin{small}
\begin{itemize}
\item \emph{``Someone in another country just logged into my account and bought three gift cards. That wasn't me!''} (\texttt{risk\_score} 0.96)\\
\textbf{\sysfour:} \texttt{BLOCK\_FRAUD} with probability 0.990; no reply.
\item \emph{``Why is my bill \$14 this month instead of \$10?''} (\texttt{invoice\_note}: ``\$4 add-on: extra storage'')\\
\textbf{\sysfour:} \reply{} with probability 0.943: ``The \$14 bill is due to the \$4 add-on for extra storage.''
\item \emph{``something is wrong with my order''}\\
\textbf{\sysfour:} \reply{} with probability 0.801: ``Could you please provide more details about the issue you're experiencing?''
\end{itemize}
\end{small}

\section{Training-data licenses}\label{app:licenses}
CLINC150: CC~BY~3.0. MASSIVE: CC~BY~4.0. Bitext: CDLA-Sharing-1.0. ReClor: non-commercial research use only. LogiQA~2.0: CC~BY-NC-SA~4.0. CosmosQA: CC~BY~4.0. PIQA: the Hugging Face card lists the licence as unknown; the official project page states the Academic Free License v3.0. $\alpha$NLI: the card lists the licence as unknown; the official repository (allenai/abductive-commonsense-reasoning) is Apache-2.0, with no separate licence stated for the data files. ProofWriter: no licence statement in the official release (its README) or on the Hugging Face copy. JevBench: MIT for the harness and the 72 original public decisions; upstream material retains its own licences \citep{jevbench2026}. Added for \sysplus{}: ABCD: MIT. Bitext verticals: CDLA-Sharing-1.0. CFPB Consumer Complaint Database: public domain (CC0 on the mirror used).

\end{document}